\pdfoutput=1

\documentclass[11pt]{article}

\usepackage[preprint]{acl}

\usepackage{times}
\usepackage{latexsym}

\usepackage[T1]{fontenc}

\usepackage[utf8]{inputenc}

\usepackage{microtype}

\usepackage{inconsolata}

\usepackage{graphicx}
\usepackage{subcaption} 
\usepackage{tcolorbox}
\tcbuselibrary{theorems}
\newtcbtheorem[]{prompt}{{ \textbf{Template to Obtain \texttt{Val\_PPL}}}}%
{colback=gray!10, colframe=gray!50!black, 
 left=0in, right=0in, bottom=0.02in, top=0.02in, fontupper=\large}{th}

\newtcbtheorem[]{promptcond}{{ \textbf{Template for Computing Conditional}}}%
{colback=gray!10, colframe=gray!50!black, 
 left=0in, right=0in, bottom=0.02in, top=0.02in, fontupper=\large}{th}

\makeatletter
\DeclareRobustCommand\onedot{\futurelet\@let@token\@onedot}
\def\@onedot{\ifx\@let@token.\else.\null\fi\xspace}

\newcommand{\eg}{\textit{e.g.}}
\newcommand{\ie}{\textit{i.e.}}

\makeatother
\usepackage{colortbl}
\usepackage{amsmath}
\usepackage{amssymb}
\usepackage{mathtools}
\usepackage{adjustbox}
\usepackage{marvosym}
\usepackage{colortbl}
\usepackage{multirow}
\usepackage{graphicx}
\usepackage{booktabs}
\usepackage{caption}
\usepackage{subcaption}
\usepackage{tabularx}
\usepackage{pifont}
\usepackage{arydshln}
\usepackage{makecell}
\usepackage{bbm}
\usepackage{wrapfig} 
\usepackage{bm}
\usepackage{algorithm}
\usepackage{algorithmic}
\usepackage{enumitem}

\usepackage{xcolor}

\definecolor{highlightcolor}{gray}{.9}

\definecolor{baselinecolor}{gray}{.9}
\definecolor{backcolor}{RGB}{232, 242, 255}

\definecolor{color1}{HTML}{1f77b4}
\definecolor{color2}{HTML}{ff7f0e}
\definecolor{color3}{HTML}{2ca02c}
\definecolor{color4}{HTML}{d62728}
\definecolor{color5}{HTML}{9467bd}

\definecolor{colora}{rgb}{0.4, 0.7607843137254902, 0.6470588235294118}
\definecolor{colorb}{rgb}{0.9882352941176471, 0.5529411764705883, 0.3843137254901961}

\def\gL{{\mathcal{L}}}

\title{Corrupted but Not Broken: Understanding and Mitigating the Negative Impacts of Corrupted Data in Visual Instruction Tuning}

\author{
 \textbf{Yunhao Gou\textsuperscript{1,2}\thanks{Equal contribution}},
 \textbf{Hansi Yang\textsuperscript{2}}$^*$,
 \textbf{Zhili Liu\textsuperscript{2,3}},
 \textbf{Kai Chen\textsuperscript{2}},
 \textbf{Yihan Zeng\textsuperscript{3}},
 \textbf{Lanqing Hong\textsuperscript{3}},
\\
 \textbf{Zhenguo Li \textsuperscript{3}},
 \textbf{Qun Liu\textsuperscript{3}},
 \textbf{Bo Han\textsuperscript{4}},
 \textbf{James T. Kwok\textsuperscript{2}},
 \textbf{Yu Zhang\textsuperscript{1}}
\\
\\
 \textsuperscript{1}Southern University of Science and Technology,
 \\
 \textsuperscript{2}The Hong Kong University of Science and Technology,
 \\
 \textsuperscript{3}Huawei Noah’s Ark Lab, \textsuperscript{4}Hong Kong Baptist University,
\\
 \small{
   \textbf{Correspondence:} \href{mailto:yu.zhang.ust@gmail.com}{yu.zhang.ust@gmail.com} 
 }
}

\begin{document}
\maketitle
\begin{abstract}
Visual Instruction Tuning (VIT) aims to enhance Multimodal Large Language Models (MLLMs), yet its effectiveness is often compromised by corrupted datasets with issues such as hallucinated content, incorrect responses, and poor OCR quality. 
Previous approaches to address these challenges have focused on refining datasets through high-quality data collection or rule-based filtering that can be costly or limited in scope. 
In this paper, we conduct a systematic investigation into the impact of corrupted data on MLLMs and discover that, although corrupted data degrade model performance, such adverse effects are largely reversible, 
and MLLMs are {\bf corrupted but not broken}. 
Specifically, we find that disabling a small subset of parameters can almost fully restore performance. Moreover, corrupted MLLMs inherently possess the capability to differentiate between clean and corrupted samples, facilitating dataset cleaning without external intervention. Building on these insights, we introduce a corruption-robust training paradigm that significantly surpasses existing strategies for mitigating the effects of corrupted data.
\end{abstract}

\begin{figure}[t]
    \centering
    \begin{minipage}[t]{\linewidth}
        \centering
        \includegraphics[width=\linewidth]{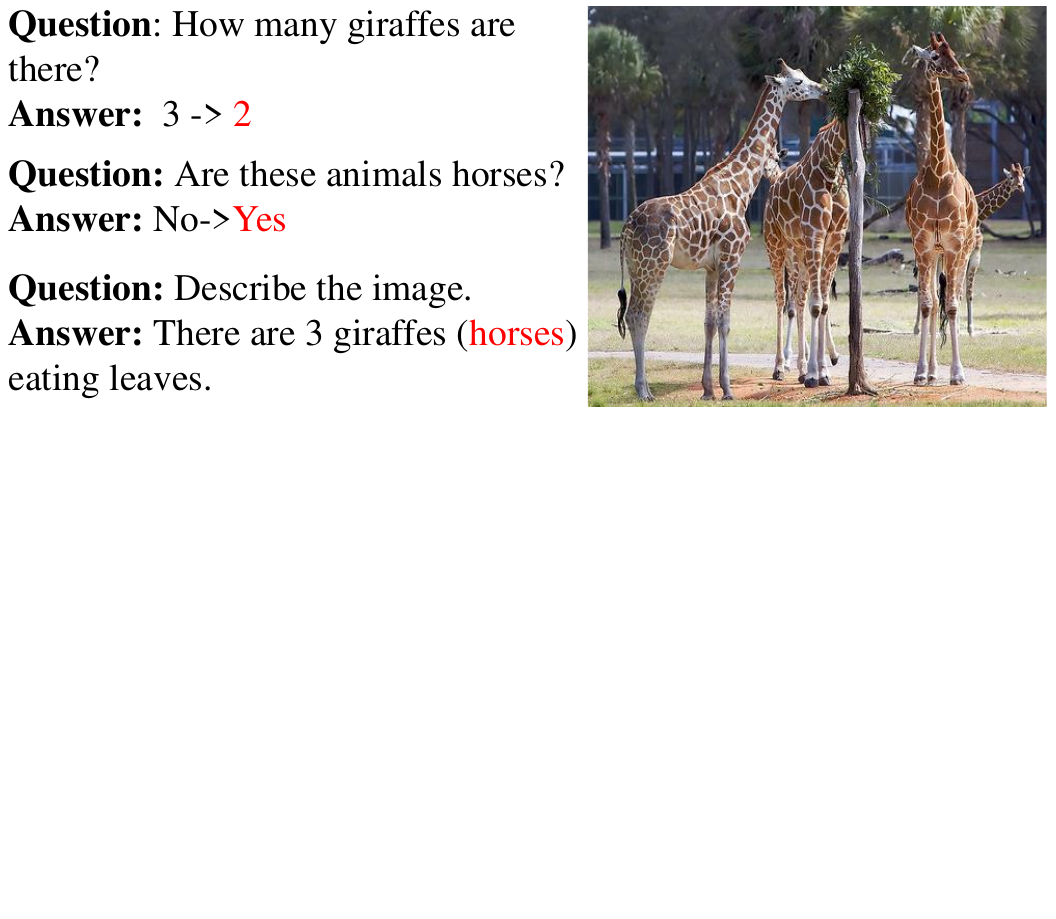}
        \caption{Examples of corrupted samples in VIT.}
        \label{fig:intro_exp}
    \end{minipage}
    \vspace{-5pt}
    \begin{minipage}[t]{\linewidth}
        \centering
        \includegraphics[width=\linewidth]{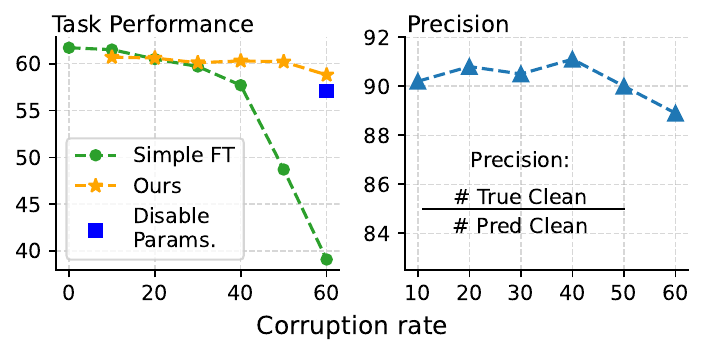}
        \vspace{-18pt}
        \caption{\textbf{Left}: Average task performance of MLLMs with various corruption ratios. Though simple fine-tuning suffers from a performance drop, disabling corruption-related parameters (1.4\%) can largely restore the performance. Our method is robust to various corruption rates. \textbf{Right}: MLLM's (fine-tuned with corrupted samples) precisions of classifying clean and corrupted samples. 
        Details are in Appendix \ref{app:detail_intro}.}
        \label{fig:intro_finding}
    \end{minipage}
\end{figure}

\vspace{-3mm}
\section{Introduction}
\vspace{-2mm}
Visual Instruction Tuning (VIT)~\citep{liu2023llava} has been actively explored to enhance the visual processing capabilities of Multimodal Large Language Models (MLLMs), extending beyond basic vision-language tasks to more complex domains such as geometric problem-solving~\citep{gao2025gllava} and chart interpretation~\citep{li2024multimodal}. To support these advancements, large-scale VIT datasets are either crawled from the Internet or synthesized using generative AI models. However, those datasets often contain corrupted data such as incorrect responses~\citep{dubey2024llama} and hallucinated content~\citep{wu2024unified} (illustrated in Figure \ref{fig:intro_exp}). 
Intuitively, 
such corruption in data should degrade the performance of MLLMs or cause abnormal behaviors. 
As such, effort has been devoted to try to mitigate the negative effect of data corruption in VIT data.
For example, LLaMA-3~\citep{dubey2024llama} conducts post-training with verified samples. Molmo~\citep{deitke2024molmo} employs human annotators to curate 1M high-quality image captions. DeepSeek-VL2~\citep{wu2024deepseek} refines responses using meta-information. However, collecting high-quality data or meta-information can be expensive. To reduce such costs, Eagle-2~\citep{li2025eagle} and Intern-VL2.5~\citep{chen2024expanding} adopt heuristic rule-based filtering. 
Such heuristic rules are derived from simple intuition
without systematic analysis on the negative effects of data corruption in VIT, 
which limits their applicability to specific types of corruptions and prevents their generalization to other scenarios.

Motivated by the limitations of existing works on the data corruption issue in VIT, in this paper, we propose to analyze the negative effects of data corruption in VIT from two perspectives:
\begin{enumerate}
    \item \emph{How does data corruption (negatively) impact the performance and behavior of MLLMs?}
    \vspace{-2mm}
    \item \emph{Do MLLMs possess any hidden capacities to overcome such negative impacts? If they do, how can we utilize them?}
\end{enumerate}

Through experiments with meticulously designed corrupted VIT datasets (details in Appendix~\ref{app:data_corrupt}-\ref{app:data_taxonomy}),
we discover the following intriguing findings:

\paragraph{The negative effect of data corruption is reversible.}     
The left part of Figure~\ref{fig:intro_finding}
shows the model performance under various corruption levels.
While corrupted 
fine-tuning
data significantly degrade the 
MLLM's 
performance,
simply disabling 1.4\% of  its
parameters 
can largely restore its performance. This suggests 
that the damage due to 
corrupted  data is restricted to only a small proportion of parameters, 
indicating that MLLMs may retain the ability on evaluation tasks even when fine-tuned on corrupted data (Section \ref{sec:re_para}).

\paragraph{Corrupted model have underlying capability to differentiate clean and corrupted samples.}
As shown in the right of Figure~\ref{fig:intro_finding}, 
MLLMs fine-tuned on corrupted data can still distinguish clean samples from corrupted ones. We provide explanations for this ability and further propose a corruption-robust training paradigm that significantly outperforms existing corruption mitigation strategies (Section \ref{sec:method}). 

In summary, our contributions are three-fold.
\begin{itemize}[leftmargin=*]
    \item 
    We are the first to systematically study the impact of corrupted data in VIT, revealing its detrimental yet reversible effect on the MLLM's performance after fine-tuning.
    \item We further demonstrate that MLLM fine-tuned with corrupted VIT data possess an underlying ability to identify clean samples in corrupted training data, 
    and we can utilize such ability to guide further fine-tuning to reverse the negative impacts of corrupted data. 
    \item 
    Empirical results across different tasks demonstrate that the underlying capabilities of the MLLM enable it to be more robust in the presence of corrupted data compared to existing approaches.
\end{itemize}


\section{Related Work}
\subsection{Data Enhancement For MLLMs}
Data corruption in VIT—such as repetitive~\citep{chen2024expanding,li2025eagle}, hallucinated responses, poor OCR quality~\citep{wu2024deepseek}, and incorrect answers~\citep{dubey2024llama}—degrades the model performance. To improve dataset quality, 
one can use
a
costly clean 
\textbf{oracle model}  to
regenerate~\citep{chen2023sharegpt4v,chen2024allava} or filter~\citep{fu2025tldr,xiong2024llava} the clean samples.
\textbf{Heuristic} methods detect corruption via patterns like repetition and image resolution~\citep{chen2024expanding,li2025eagle} but fail to address hallucinations and incorrect responses. In addition, the lack of a comprehensive understanding of how corruption affects MLLMs limits the development of more effective mitigation strategies. Our work fills this gap by analyzing the impact of corrupted samples and proposing a robust solution.

\subsection{Learning with Noisy Labels (LNL)}
Our study is related to learning with noisy labels (LNL) in machine learning, which aims to mitigate the effect of mis-labeled data when training a classification model. It can be generally categorized into three main approaches~\citep{han2020survey}:
designing special loss functions that are robust to possibly wrong supervision~\cite{ghosh2017robust,zhang2018generalized,menon2019can}, 
correcting wrong supervision with model prediction~\citep{tanaka2018joint,yi2019probabilistic,zhang2020distilling},
and sample selection, which identifies noisy samples from the training data and then makes them less influential in the training process~\cite{jiang2018mentornet, han2018co,wei2020combating}. 
Among these approaches, sample selection based on the memorization effect~\cite{arpit2017closer,zhang2016understanding},
which considers samples with small loss values as clean samples~\citep{han2018co,jiang2018mentornet,yao2020searching,10509799}, 
usually achieves the best performance.  
However, these approaches cannot effectively leverage the instruction following abilities of MLLMs, 
which prevents them to effectively handle corrupted VIT data.


\begin{figure*}[!]
\centering
\includegraphics[width=\linewidth]{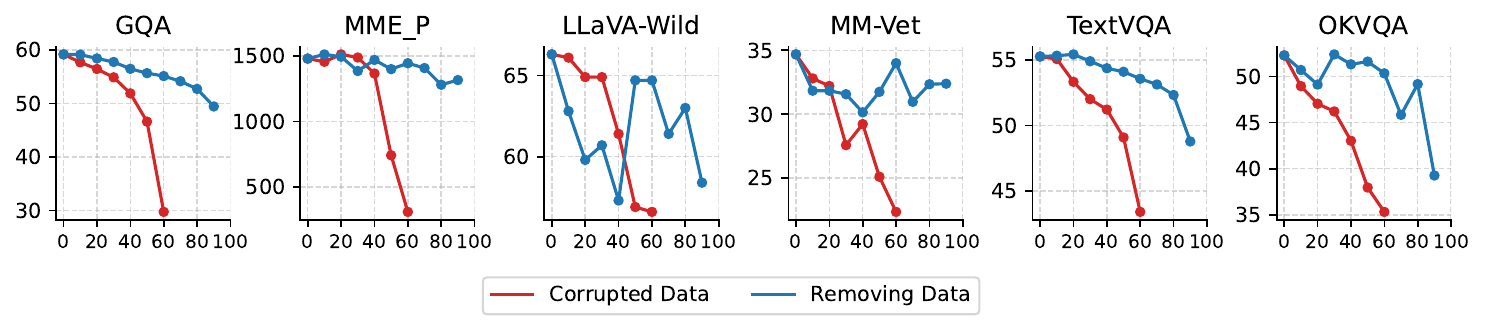}
\caption{\textbf{Performance (y-axis) of LLaVA-1.5 (LLaMA-3.1-8B) under different corruption ratios (x-axis).}}
\label{fig:effect_nr_main}
\end{figure*}

\section{Preliminaries}
\label{sec:preliminary}

\paragraph{Notations.}
Given an image \( x_v \) and the corresponding instruction \( x_q \),
an MLLM (parameterized by \( \bm{\theta} \)) predicts a response \( \hat{x}_y \).
For simplicity of notations, let \( x_c \equiv (x_v, x_q) \). The fine-tuning process optimizes the model on a dataset \( \mathcal{D} = \{(x_c, x_y)\} \)
by minimizing the loss \( \ell({x}_y |x_c;\bm{\theta}) = -\log p(x_y | x_c; \bm{\theta}) \).  For convenience, we sometimes use \( x\) to denote \( (x_c, x_y) \) and simplify the loss 
$\ell(x_y | x_c; \bm{\theta})$
as $\ell(x; \bm{\theta})$.

\paragraph{Corruption.} In this paper, we mainly focus on 
image-text alignment
corruption,
\ie, 
given an image and the instruction, the response can be incorrect. This covers common corruptions in VIT such as incorrect and hallucinated responses. On the other hand, text-only corruptions, such as grammar errors and repetitions, are not the focus of this paper.

To construct 
image-text alignment
corruptions, we 
replace the correct responses with 
incorrect 
ones
generated
from GPT-4o~\footnote{https://genai.ust.hk/}.
Specifically, let $z \in \{0,1\}$ be the correctness of a sample \(x_c\). A dataset with corruptions is $\mathcal{\Tilde{D}} = \left\{ (x_c, \tilde{x}_y) \right\}$, with 
$$\tilde{x}_y = \begin{cases} 
x_y & \text{if } z = 1 \ \text{(clean)}\\
g(x_c, x_y) & \text{if } z = 0 \  \text{(corrupted)}
\end{cases},$$
where $g$ denotes GPT-4o. The prompt used and examples of corrupted data are shown in Appendix~\ref{app:data_corrupt}.
We define the corruption ratio $cr$ as the proportion of corrupted data (\ie, those with $z=0$) in $\Tilde{\mathcal{D}}$.


\paragraph{Experimental Setup.}

We follow the setup of LLaVA-1.5 \citep{liu2023improved}. Specifically, we begin by pre-training the vision projectors, which connect the CLIP visual encoder ViT-L/14 \citep{radford2021learningtransferablevisualmodels} to the LLM (\eg, LLaMA-3.1-8B~\citep{dubey2024llama} and Qwen-2.5 0.5B/3B/7B models~\citep{qwen2.5}), using approximately 600K image-text caption pairs. Then, we perform supervised fine-tuning (SFT) with an 100K instruction-tuning dataset, which is uniformly sampled from LLaVA-665K~\citep{liu2023improved} with 665K text-only and vision-language instances. Note that for all experiments throughout this paper, we only inject corruption into the instruction-tuning dataset with images, excluding pre-training dataset and text-only instruction-tuning datasets. 

Following LLaVA-1.5, we evaluate the performance 
on 11 standard evaluation datasets. Based on the response formatting prompts, all the training and evaluation datasets are 
divided into the following groups:
(i) VQA (visual question answering);
(ii) MC-VQA (multiple-choice VQA); and
(iii) Conversation. Note these groups only differ in response formatting groups, \ie, they share the same underlying vision-language content.
More details on the 
datasets and performance metrics are in Appendices \ref{app:data_taxonomy} and \ref{app:data_metrics}, respectively.

\begin{figure*}[!]
\centering
\includegraphics[width=\linewidth]{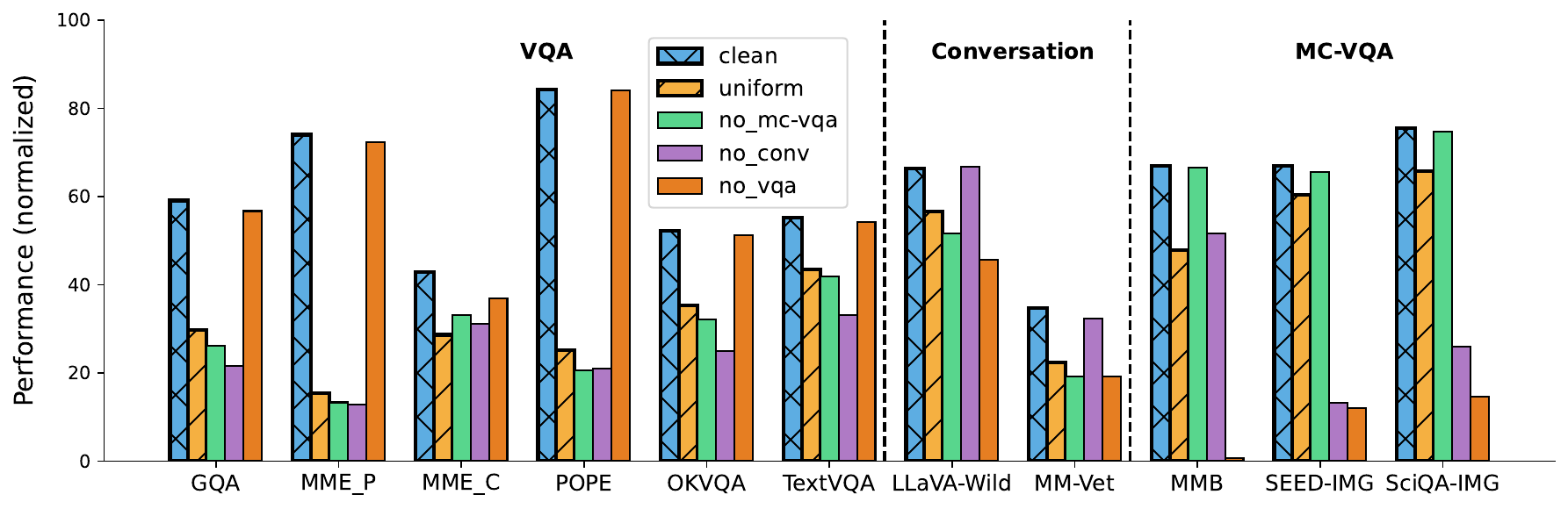}
\caption{\textbf{Effects of corruption on LLaVA-1.5 (LLaMA-3.1-8B).} The evaluation datasets are shown in 3 groups: VQA, Conversation and MC-VQA. The corruption ratio here is 60\%. }
\label{fig:effect_nr_task}
\end{figure*}

\section{Effects of Data Corruption in VIT}
\label{sec:effect}

To analyze how data corruption may impact the MLLM's performance on downstream tasks, 
we first consider the simplest way of introducing data corruption 
that uniformly drawing clean samples from all datasets and replacing them with corrupted ones.
Denote the ratio of corrupted samples in the whole dataset as $cr$ (corruption ratio). 
We vary $cr$ from 0\% to 60\% and further construct a reference dataset with those corrupted samples removed to see whether they are contributing negatively. 
Figure~\ref{fig:effect_nr_main} shows the performance of MLLM fine-tuned under different corruption ratios on various benchmarks (results on more datasets are in Appendix \ref{app:more_effect}). 
For most tasks, corrupting the data results in worse performance than simply removing them. This degradation worsens with increasing $cr$, indicating the negative effect of corrupted data.

While such negative effect well matches our intuition, our further investigation will reveal that such negative effect is \textit{restricted} and largely \textit{reversible}. 
In Section~\ref{sec:effect_benchmark}, we demonstrate that data corruption only on specific tasks does not yield negative effects on other tasks. 
In Section~\ref{sec:effect_parameter}, we further demonstrate that such effect can be largely reverted by 
removing affected parameters that only take a small proportion of the whole MLLM and that MLLMs fine-tuned with corrupted data retain capabilities on the evaluated tasks.


\subsection{Negative Effects of Data Corruption is Restricted to Tasks with Corrupted Data}
\label{sec:effect_benchmark}

{\bf }

Despite uniformly selecting clean samples from all datasets and replacing them with corrupted ones, we further consider another type of corruption that 
only selectively corrupts clean samples for specific tasks:
(i) \texttt{no\_vqa}: 
corruption is injected into all datasets except the VQA datasets; (ii)
\texttt{no\_mc-vqa} in which the corruption is injected into all datasets except the multiple-choice VQA datasets; 
and (iii) 
\texttt{no\_conv} in which the corruption is injected into all datasets except the conversation datasets. 

Figure~\ref{fig:effect_nr_task} compares the
performance 
of MLLMs fine-tuned on datasets
with different types of corruption. 
While Figure~\ref{fig:effect_nr_main} demonstrates that data corruption leads to performance degradation, 
{\it such effect is limited to tasks that contain corrupted training data and will not extend to other tasks without corruption. 
}
For example, on the VQA task, the MLLMs fine-tuned on \texttt{no\_vqa} perform comparably to those fine-tuned on \texttt{clean} and considerably better than those fine-tuned on \texttt{uniform} across different datasets. 
This observation remains valid for models fine-tuned on the \texttt{no\_mc-vqa} and \texttt{no\_conv} datasets, when evaluated on the multiple-choice VQA and conversation tasks, respectively.
In other words, even if the majority of training data are corrupted ($cr=60\%$), the model can still maintain its performance on specific tasks as long as the corresponding training data are not corrupted. 

\paragraph{Implications.} Note that as detailed in Appendix \ref{app:data_taxonomy}, the VQA, MC-VQA and Conversations tasks experimented in this section differ only in their response prompt format rather than the underlying vision-language (or multi-modal) knowledge. Therefore, the restricted effect of corrupted VIT data revealed in this section indicates that even the MLLM is corrupted, it still retains recoverable capabilities on the evaluated tasks. We shall discuss detailed recovery strategies in the next section.

\subsection{Negative Effects of Data Corruption is Reversible}
\label{sec:effect_parameter}

\begin{table}
    \centering
     \resizebox{0.48\textwidth}{!}{
        \begin{tabular}{lccc}
        \toprule
        Model               & Avg. Score & \% Disabled & ($p$, $q$)        \\ \midrule
        \texttt{Clean}               & 61.7      & -          & -             \\ 
        \texttt{Clean (40K)}         & 59.3      & -          & -             \\ \midrule
        \multirow{4}{*}{\setlength\extrarowheight{0pt}\begin{tabular}[l]{@{}l@{}}\texttt{Corrupted}\\ \texttt{($cr=60\%$, 100K)}\end{tabular}}      & 39.1      & 0          & -             \\ 
                                    & 51.4      & 0.84     & (12, 10)      \\ 
                                    & 55.2      & 1.16     & (17, 15)      \\ 
                                    & 57.1      & 1.39     & (22, 20)      \\ \bottomrule

        \end{tabular}%
        }
              \caption{\textbf{
        Performance of LLaVA-1.5 (LLaMA-3.1-8B) with corruption-related weights disabled}.
        }
        \label{tab:param}
        
\end{table}

\begin{figure*}[t]
\centering

    \begin{subfigure}[b]{\textwidth}
        \centering
        \includegraphics[width=\linewidth]{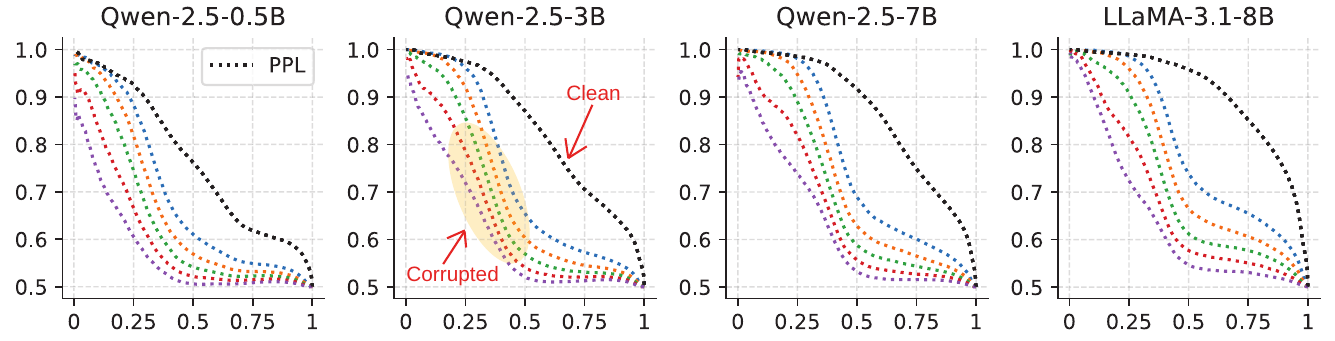} %
        \caption{\textbf{\texttt{PPL}}}
        \label{fig:precision_50_qwen_ppl}
    \end{subfigure}
    \hfill
    \begin{subfigure}[b]{\textwidth}
        \centering
        \includegraphics[width=\linewidth]{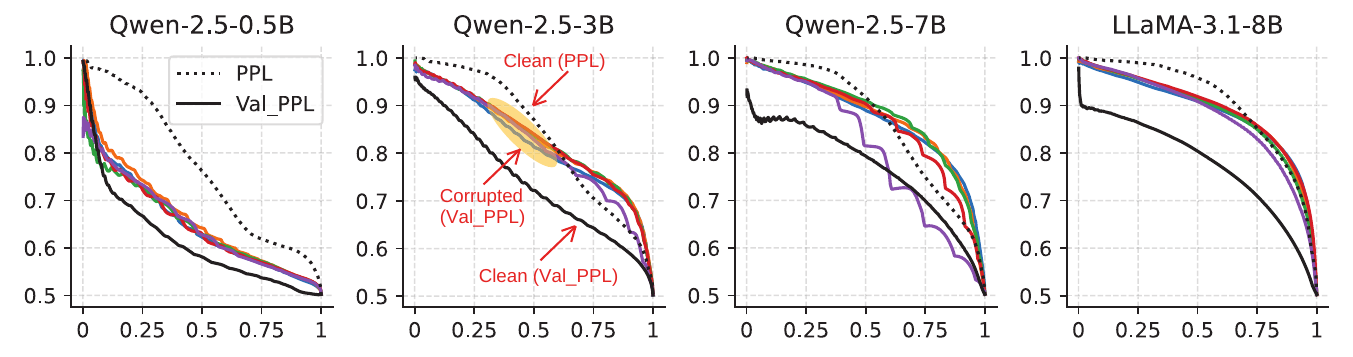} %
        \caption{\textbf{\texttt{PPL} and \texttt{Val\_PPL}}}
        \label{fig:precision_50_qwen_ppl_val_ppl}
    \end{subfigure}

\caption{\textbf{Precision-recall curves of MLLM's predictions on the correctness of 100K samples ($cr=50\%$).} $x$-axis: recall; $y$-axis: precision. Solid and dotted line denote predictions based on \texttt{Val\_PPL} and \texttt{PPL}, respectively.  Color represents the corruption ratio of the training dataset: \textcolor{black}{\Large{$\bullet$}}\textcolor{black}{0\%}, \textcolor{color1}{\Large{$\bullet$}}\textcolor{color1}{10\%}, \textcolor{color2}{\Large{$\bullet$}}\textcolor{color2}{20\%} \textcolor{color3}{\Large{$\bullet$}}\textcolor{color3}{30\%}, \textcolor{color4}{\Large{$\bullet$}}\textcolor{color4}{40\%}, \textcolor{color5}{\Large{$\bullet$}}\textcolor{color5}{50\%}. }
\label{fig:precision_50_qwen}
\end{figure*}


\label{sec:re_para}

To recover the performance of the MLLM fine-tuned on corrupted data, we remove the model parameters affected by corrupted data.
To achieve this goal, we first try to identify weights in the MLLM that are particularly responsible for generating corrupted responses. 
Specifically, following \citet{wei2024assessing}, we select weights with top-\(q\%\) influence scores on a \textbf{corrupted} dataset
but remove those that overlap with weights with top-\(p\%\) influence on a \textbf{clean} dataset.
This ensures that only weights contributing specifically to corrupted samples are considered. We use the SNIP score ~\citep{lee2018snip} to compute the influence,
and consider an MLLM fine-tuned on 100K VIT data samples
with corruption ratio \(cr=60\%\). 
With different choices 
of $(p,q)$, we remove weights that are only related to the corrupted data 
and report the performance after removing these weights. For comparison, we include results from 
models fine-tuned with 100K and 40K clean VIT data samples
(denoted by \texttt{Clean} and \texttt{Clean(40K)},
respectively). See more details on in Appendix \ref{app:prune}.

Table~\ref{tab:param}
shows the performance on the evaluation tasks
after removing weights related to corrupted data.
By removing fewer than \textbf{1.4\%} of the parameters, the corrupted model 
can restore its performance from 39.1 to 57.1, which is already close to 
that
(\ie, 59.3) 
of the model fine-tuned with 40K clean data (all clean data under a $cr$ of 60\%). 

\paragraph{Implications.} Notably, removing weights relevant to corrupted data aims to ``delete'' effects of corrupted data rather than ``override'' with new knowledge. This confirms that the corrupted MLLM does retain capabilities on evaluated tasks. 

\section{Underlying Capabilities of MLLMs Fine-tuned with Corrupted Data}
\label{sec:method} 

Motivated by the restrictive and reversible nature of the negative impact of corrupted data as analyzed in Sections~\ref{sec:effect_benchmark} and~\ref{sec:effect_parameter}, 
we conjecture that the MLLMs fine-tuned with corrupted data may possess the underlying capacity to identify clean training samples and introduce a simple method called \textbf{self-validation} in Section~\ref{sec:filter}. 

By further fine-tuning corrupted MLLMs with samples selected by the proposed self-validation process, we demonstrate that the corrupted MLLM can recover from the negative impacts of corruption without any external annotations in Section~\ref{sec:final_method}.

\subsection{Corrupted MLLM Can Detect Clean Samples from Corrupted Dataset}
\label{sec:filter}

To identify clean and noisy samples from a corrupted VIT dataset, 
a straight-forward idea following existing works on learning with label noise is to consider the sample loss~\cite{jiang2018mentornet,han2018co}: 
larger sample loss may indicate the corresponding sample is likely to contain wrong responses. 
That can be implemented using the perplexity score~\citep{marion2023less},
as we can prove that it is proportional to the MLLM's training loss (details in Appendix \ref{app:ppl}). We term this score as \texttt{PPL} in this paper.

Figure~\ref{fig:precision_50_qwen_ppl} compares the recalls and precisions (see details in Appendix \ref{app:classify}) of using \texttt{PPL} from MLLM fine-tuned with varying corruption levels
to identify clean and noisy samples. 
As can be seen, MLLMs fine-tuned with clean data (the black dotted line) can accurately identify clean samples by only keeping small-loss samples. For example, Qwen-2.5-3B achieves over 0.85 precision at a recall of 0.5. On the contrary, MLLMs fine-tuned with corrupted data (non-black dotted lines) achieve worse performance with increasing corruption levels. Specifically, all their precisions drop below 0.65 at a recall of 0.5. 
\emph{In other words, existing loss-based sample selection approaches are not robust to data corruption 
and cannot be simply extended to VIT.}

\subsubsection{Utilize Self-Validation to Activate Corrupted MLLM's Capability}

Given that the negative effect of corrupted data is restricted to specific response formats (Section \ref{sec:effect_benchmark}) and that the corrupted MLLMs still retain capabilities on the evaluated tasks (Section \ref{sec:effect_parameter}),
we propose to instruct the corrupted MLLM with different response format that is unseen during training. 
Specifically, 
using the following template,
we directly prompt the MLLM to predict whether a sample is corrupted:
    \begin{prompt*}{}{}
{\textit{<image>Query:} $\left\{\text{instruction text}\right\}$ \\
\textit{Response:} $\left\{\text{response text}\right\}$ \\
\textit{Is the response correct? Answer yes or no:}
}
\end{prompt*}
The perplexity of the predicted word “No” 
is then used
as the score. 
To distinguish from the perplexity score of responses, 
we refer the perplexity of word ``No'' in self-validation as \texttt{Val\_PPL}.
Contrary to \texttt{PPL}, samples with smaller \texttt{Val\_PPL} score are considered more likely to be corrupted. 

\paragraph{Self-Validation is robust against corrupted data.} Figure~\ref{fig:precision_50_qwen_ppl_val_ppl} 
compares the recalls and precisions of using the  \texttt{Val\_PPL} scores (solid curves) from MLLM fine-tuned with varying corruption levels 
to identify clean and corrupted samples. 
For reference, we also include the corresponding precision-recall curve of clean MLLMs
previously shown in Figure~\ref{fig:precision_50_qwen_ppl} (black dotted curve). 
Remarkably, for MLLMs with 3B parameters and above, models fine-tuned with corrupted data (non-black solid lines) obtain similar precision-recall curves compared to that of the clean model using \texttt{PPL}. Further, even at increasing corruption levels, those curves are consistent and close to each other (except for Qwen-2.5-7B at $cr$ of 50\%). This demonstrates that corrupted MLLMs can effectively distinguish clean and corrupted samples using \texttt{Val\_PPL}.

However, we note that the advantage
of \texttt{Val\_PPL} 
is less significant for the small-sized LLM (Qwen-2.5-0.5B).
This suggests that self-validation is an emergent ability that is only observed in larger language models.

\begin{figure}[t]
\centering
\includegraphics[width=\linewidth]{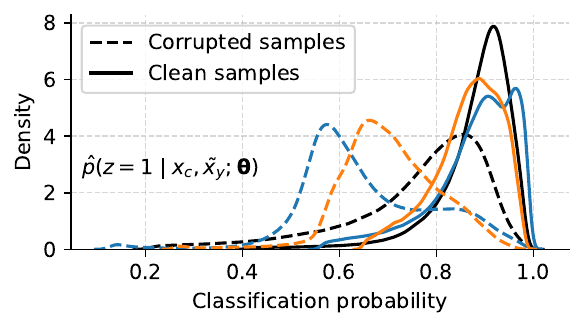}
\caption{Distribution of classification probability
\( \hat{p}(z=1|x_c, \tilde{x}_y; \bm{\theta}) \).
Color represents corruption ratios of datasets the model is trained on: \textcolor{black}{\Large{$\bullet$}}\textcolor{black}{0\%},\textcolor{color1}{\Large{$\bullet$}}\textcolor{color1}{10\%}, \textcolor{color2}{\Large{$\bullet$}}\textcolor{color2}{20\%}.}
\label{fig:posterior}

\end{figure}

\subsubsection{Why do Corrupted Data Improve the Performance of Self-Validation?}
\label{sec:filter_analysis}
An intriguing observation from
Figure \ref{fig:precision_50_qwen_ppl_val_ppl} is that
with self-validation, the corrupted model can identify corrupted samples
even better than the clean model, 
as the solid black curves are always below the solid non-black curves. 
In this section,
we provide an empirical analysis for this phenomenon.

\paragraph{Probabilistic modeling of self-validation.}
Recall that
the \texttt{Val\_PPL} template asks the MLLM to predict correctness of a sample. Essentially, the MLLM estimates 
\( p(z=1|x_c, \tilde{x}_y) \), the 
(ground-truth)
probability that response $\tilde{x}_y$ is correct,
with its output probability 
\( \hat{p}(z=1|x_c, \tilde{x}_y; \bm{\theta}) \).
Figure~\ref{fig:posterior}
shows the distributions of 
\( \hat{p}\) on the clean samples and corrupted samples
from a corrupted VIT dataset 
that is also used to finetune the MLLM. 
We can see that
corruption in the fine-tuning data 
has little effect on the  output probability distribution
for clean samples (solid curves). This results in robustness of \texttt{Val\_PPL} at various corruption ratios and can be explained by the restrictive nature of corrupted data on MLLMs as the corrupted dataset does not contain self-validation instructions. 
In contrast,
significant distribution 
shift is observed on that of the corrupted samples (dashed curves), which is the cause of improved performance of \texttt{Val\_PPL} when fine-tuned with corrupted data. However, this cannot be explained by the conclusions drawn so far.

\begin{table*}[h!]
\centering
\resizebox{\textwidth}{!}{%
\begin{tabular}{lcccccccccccc}
\toprule
Methods & \textbf{Avg.} & GQA & MME\_P & MME\_C & POPE & \setlength\extrarowheight{0pt}\begin{tabular}[c]{@{}c@{}}LLaVA\\ Wild\end{tabular} & MM-Vet & MMB & \setlength\extrarowheight{0pt}\begin{tabular}[c]{@{}c@{}}SEED\\ IMG\end{tabular}  & \setlength\extrarowheight{0pt}\begin{tabular}[c]{@{}c@{}}SciQA\\ IMG\end{tabular}  &  \setlength\extrarowheight{0pt}\begin{tabular}[c]{@{}c@{}}Text\\ VQA\end{tabular} & OKVQA \\ \midrule

\rowcolor{baselinecolor}
Clean & 61.65 & 59.18 & 1480.36 & 342.86 & 84.25 & 66.30 & 34.68 & 66.92 & 66.91 & 75.51 & 55.25 & 52.28  \\ 
None (CE) & 49.13  & 41.87  & 668.17  & 253.21  & 62.90  & 57.30  & 23.47  & 63.83  & 63.26  & 74.02  & 50.01  & 38.67   \\ \midrule
\multicolumn{12}{l}{\textit{Noise-robust loss functions}} \\
GCE & 50.95  & 40.67  & 751.27  & 240.00  & 69.37  & 62.00  & 23.85  & \underline{65.81}  & \underline{64.84}  & \textbf{74.81}  & 50.05  & \underline{41.51}   \\ 
Phuber CE & 47.28  & 37.24  & 595.69  & 258.21  & 46.77  & 59.90  & 26.93  & 61.51  & 63.49  & 73.82  & 48.24  & 40.16   \\ \midrule
\multicolumn{12}{l}{\textit{Sample selection}} \\
MentorNet & 46.21  & 40.07  & 746.12  & 261.43  & 69.67  & 60.20  & 27.84  & 46.13  & 49.39  & 62.82  & 47.57  & 34.69   \\ 
Co-teaching & 47.97  & 39.95  & 583.35  & 253.93  & 66.38  & 57.60  & \underline{29.63}  & 55.58  & 60.91  & 72.14  & 48.83  & 35.68   \\ 
JoCoR & 47.00  & 39.23  & 571.96  & 245.36  & 59.09  & 58.40  & 27.80  & 55.33  & 60.28  & 72.19  & 48.69  & 36.76   \\ \midrule
\multicolumn{12}{l}{\textit{Further fine-tuning}} \\
EL2N & 47.07  & 49.34  & \underline{1357.43}  & 269.64  & \textbf{83.97}  & 58.20  & 27.34  & 12.97  & 43.49  & 51.96  & \underline{50.62}  & 38.27   \\ 
GradNorm & \underline{55.77}  & \underline{50.06}  & 1342.40  & \textbf{318.93}  & 76.86  & \underline{66.50}  & 27.25  & 59.79  & 64.01  & 73.43  & 49.07  & 39.48   \\ 
Entropy & 48.60  & 43.76  & 1118.39  & 261.79  & 73.53  & 60.30  & 27.57  & 42.10  & 52.68  & 63.01  & 48.12  & 34.94   \\ 
PPL & 54.69  & 47.54  & 1222.56  & 279.64  & \underline{82.65}  & 60.50  & 25.69  & 64.86  & 62.83  & 73.38  & 49.04  & 39.02   \\ 
\rowcolor{backcolor}
Val\_PPL & \textbf{60.17}  & \textbf{56.65}  & \textbf{1510.48}  & \underline{297.50}  & 82.18  & \textbf{69.30}  & \textbf{31.51}  & \textbf{67.18}  & \textbf{65.48}  & \underline{74.62}  & \textbf{53.48}  & \textbf{48.76}   \\ 
\bottomrule
\end{tabular}%
}
\caption{\textbf{Comparisons of different corruption-robust strategies at a corruption ratio of 50\% on LLaVA-1.5 (LLaMA-3.1-8B)}, where \textbf{Avg.} refers to the average results on 11 benchmarks (normalized to 0-100). Best results are in \textbf{bold}, and the second best are \underline{underlined}. Results for Qwen-2.5 series are in Tables \ref{tab:main_qwen_0.5b}, \ref{tab:main_qwen_3b}, and \ref{tab:main_qwen_7b} in the appendix.}
\label{tab:main_llama3}
\vspace{-10pt}
\end{table*}



\paragraph{Improved understanding of corrupted data} leads to distribution shift of \( \hat{p}(z=1|x_c, \tilde{x}_y; \bm{\theta}) \) on the corrupted samples (see Appendix \ref{app:analysis_improved} for a detailed analysis). Although there are no explicit annotations indicating which samples are corrupted, the MLLM is still, to some extent, aware of the corrupted data during fine-tuning. This awareness stems from its prior pre-training on image-text pairs, which enables the model to learn correct image-text alignment.
However, the \texttt{PPL} score cannot benefit from the improved understanding of corrupted data because it shares the same response format as the corrupted data and is thereby affected.

\begin{figure*}[!]
\centering
\includegraphics[width=\linewidth]{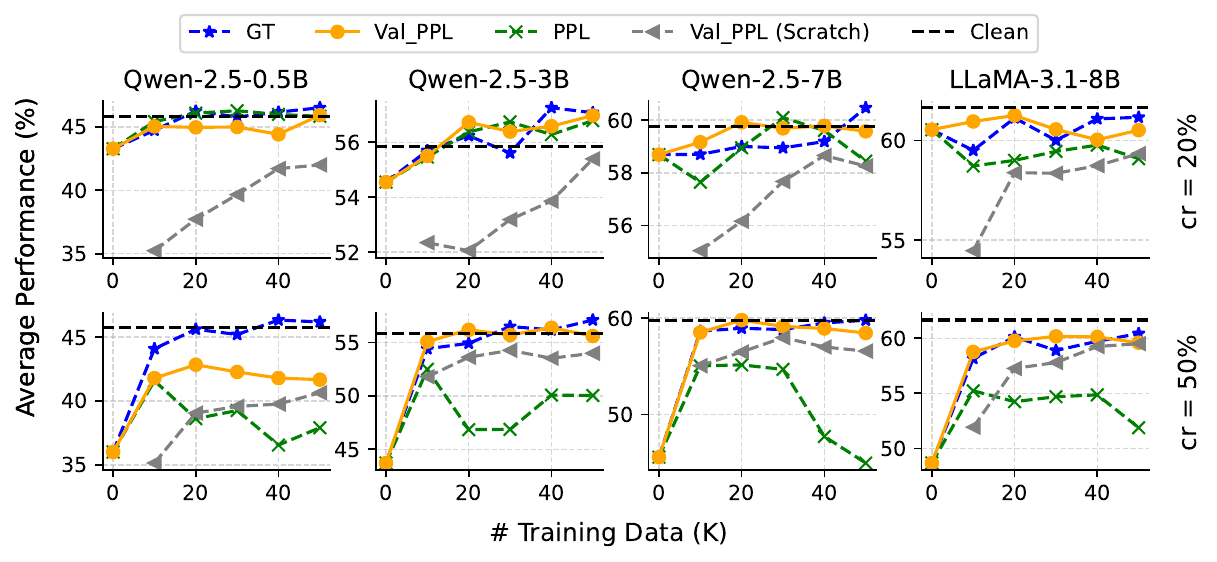}
\caption{\textbf{Average task performance of models further fine-tuned (or fine-tuned from scratch) on data from different sources}. The corruption ratios are 20\% (low) and 50\% (high), results for other corruption levels are in Figure~\ref{fig:ablate_more} in the appendix.} 
\label{fig:ablate}
\vspace{-10pt}
\end{figure*}

\subsection{Corrupted MLLM can Recover from Data Corruption by Self-Validation}
\label{sec:final_method}

To further assess the quality of the samples selected from self-validation with the \texttt{Val\_PPL} score, 
we investigate whether these samples can be used to recover the benchmark performance of MLLM previously fine-tuned on corrupted data.
We conduct experiments on the 11 benchmark datasets introduced in Section~\ref{sec:preliminary}
with the corruption ratio $cr=50\%$, by using LLaVA-1.5 built on LLama-3.1-8B and Qwen-2.5 series models.
Specifically, we further fine-tune the MLLMs previously trained on corrupted data with the samples selected by from self-validation. By default, we choose samples with the highest 30\% \texttt{Val\_PPL} scores for further fine-tuning (\ie, 30K).
In addition to using samples selected from self-validation with the \texttt{Val\_PPL} score, we also consider using the following scores to select samples: (i) the \texttt{PPL} score proposed in Section~\ref{sec:filter}; (ii) \texttt{EL2N} and \texttt{GradNorm}, which measure the $\ell_2$-norm of the output error vector and the gradient, respectively~\citep{paul2021deep}; and (iii) \texttt{Entropy}~\citep{Coleman2020Selection}, which reflects uncertainty in the output probabilities. See detailed formulae of these scores in Appendix \ref{app:sample_select_further}.

To better benchmark the performance of using self-validation for performance recovery, 
we also introduce the following baseline methods from traditional LNL (learning with noisy labels) methods:
\begin{enumerate}[leftmargin=*]
\item Noise-robust loss functions.
By default, the MLLM is trained with the Cross-Entropy (CE) loss, denoted by \texttt{None (CE)}. We consider two noise-robust loss functions from \citet{menon2019can}: (i) Generalized Cross-Entropy (\texttt{GCE}) loss~\citep{zhang2018generalized}, that combines CE loss and 
mean absolute error (MAE),
which is more robust to label noise~\citep{ghosh2017robust}, 
and 
(ii) the Phuber Cross-Entropy loss~\citep{menon2019can} (\texttt{Phuber CE}), which incorporates gradient clipping into CE. Their detailed formulae are in Appendix \ref{app:loss_func}.

\item Online sample selection methods,
which focus on selecting clean samples during training. \texttt{MentorNet} \citep{jiang2018mentornet} identifies small-loss samples as clean. \texttt{Co-teaching} \citep{han2018co} trains two networks and exchanges small-loss samples between them to avoid error accumulation. \texttt{JoCoR}~\citep{wei2020combating} enforces agreement between networks to prevent biased selection. Their details are in Appendix~\ref{app:sample_select}.

\end{enumerate}

\paragraph{Results.} Table~\ref{tab:main_llama3} 
compares the performance of various corruption-robust strategies. 
By using the \texttt{Val\_PPL} score to select clean VIT data samples for further fine-tuning, 
we can significantly restore the performance of a corrupted model, improving it from 49.13 to 60.2 on average, where the clean model achieves 61.65. 
We also outperform all existing baseline methods on average and achieves the best results on 8 out of 11 evaluation tasks, 
which further justify the high quality of VIT data samples selected by the \texttt{Val\_PPL} score.

\subsubsection{Analysis}
\label{sec:ablation}
To further assess the effectiveness of self-validation, we conduct further fine-tuning on LLaVA-1.5 fine-tuned with 20\% and 50\% corruption ratios, which cover low and high corruption in practical scenarios,\footnote{Results for other corruption levels are in Figure~\ref{fig:ablate_more} of the appendix.} using ``clean data" from various sources. (i) \texttt{GT}: Clean data with responses directly from the clean dataset.
This can be regarded as the ``best" clean data possible);
(ii) \texttt{PPL} (we choose it rather than GradNorm or Entropy because it is widely adopted in LNL and we studied it in Section \ref{sec:filter});
and (iii) the proposed \texttt{Val\_PPL}. In addition, we also experiment with 
(iv) \texttt{Val\_PPL(Scratch)}, which fine-tunes the model from scratch using the samples selected by \texttt{Val\_PPL}
rather than after it is fine-tuned on corrupted data. 

Figure~\ref{fig:ablate} shows
the average task performance of these models further fine-tuned (or fine-tuned from scratch) 
using ``clean data'' from various sources with different sizes.
As can be seen, with \texttt{GT}, most MLLMs could largely restore the performance using 20K samples, which indicates that given clean data, further fine-tuning can be a fast and cost-effective approach for performance recovery on corrupted MLLMs. Notably, both at low (20\%) and high (50\%) corruption ratios, \texttt{Val\_PPL} performs as well as \texttt{GT} for the 3B and larger MLLMs, rapidly approaching the performance of the clean model. However, for Qwen-2.5-0.5B, only \texttt{GT} can restore the model performance. This is consistent with our observation in Figure \ref{fig:precision_50_qwen_ppl_val_ppl} that the self-validation is an emergent ability for larger LLMs. On the contrary, though \texttt{PPL} could achieve comparable results with to \texttt{Val\_PPL} at low corruption, it is less effective at high corruption, demonstrating the robustness of self-validation at various corruption ratios. 
Further, we find that \texttt{Val\_PPL(Scratch)} is consistently outperformed by its further fine-tuning counterparts, demonstrating the sample-efficiency of further finetuning.

\vspace{-5pt}
\section{Conclusions}
In summary, we show that the negative impact of corrupted data on MLLMs is largely reversible and that these models can inherently identify corrupted samples. Leveraging these insights, we introduce a corruption-robust training method that outperforms existing strategies, improving the robustness of Visual Instruction Tuning.

\section*{Limitations}
One limitation of this paper is that we did not study MLLMs with larger LLMs (\eg, 70B and 400B) due to limited computational resources. 

\section*{Ethic Statement}
There is no ethical problem in our study.

\bibliography{custom}


\appendix

\renewcommand*\contentsname{Appendix}
\clearpage

\setcounter{tocdepth}{-1}  

\tableofcontents  

\addtocontents{toc}{\protect\setcounter{tocdepth}{2}}

\section{Details in Figure \ref{fig:intro_finding}}
\label{app:detail_intro}
\begin{figure*}[!]
\centering
\includegraphics[width=\linewidth]{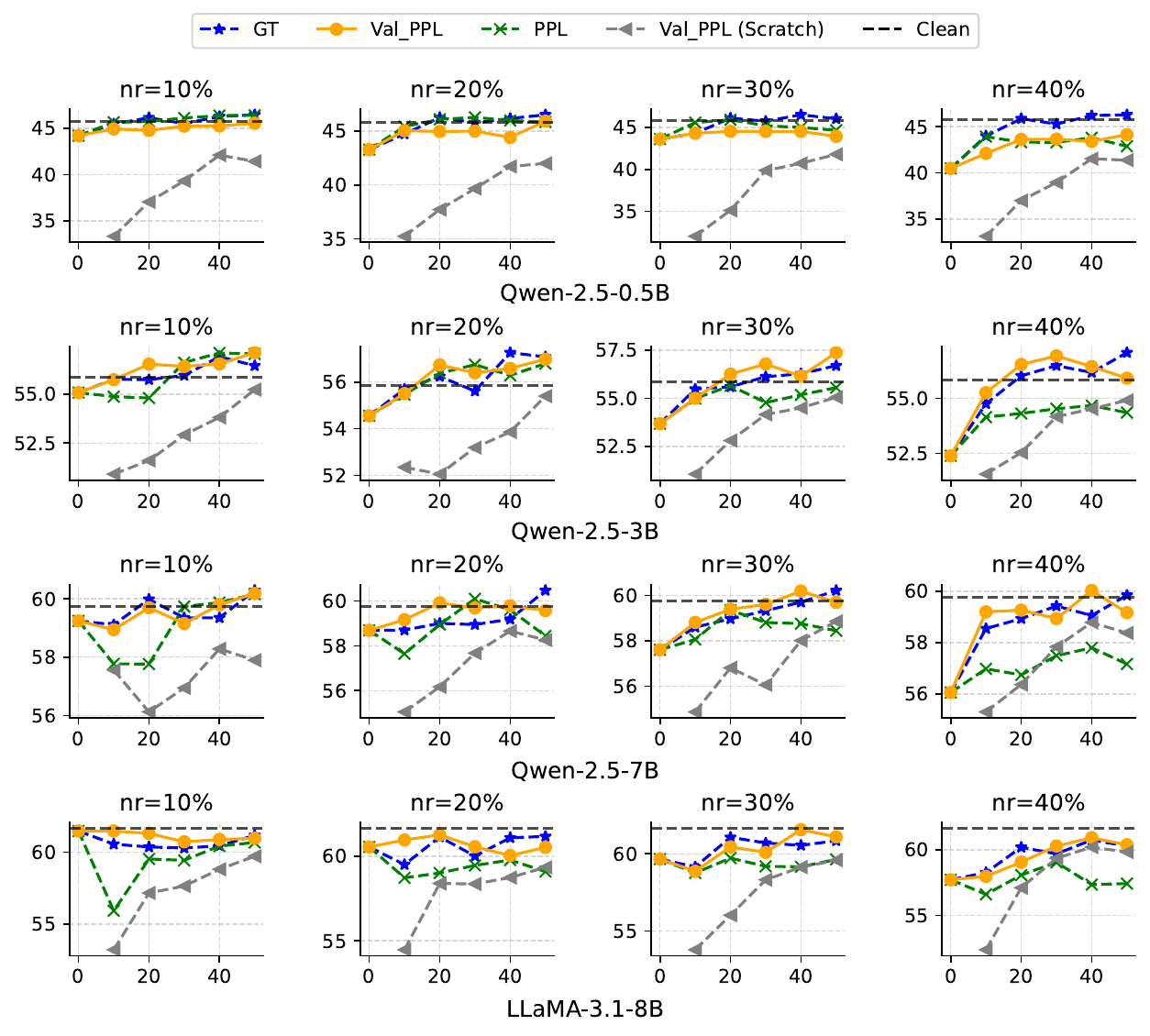}
\vspace{-.3in}
\caption{\textbf{Average model performance of models further fine-tuned (or fine-tuned from scratch) on data from data from different sources}. Further fine-tuned models are trained on dataset with various corruption levels (10\%-40\%) at first. Results with 20\% and 50\% corruption are in Figure~\ref{fig:ablate}.}
\label{fig:ablate_more}
\end{figure*}

The ``Simple FT'' results in Figure \ref{fig:intro_finding} are aggregated from Figure \ref{fig:effect_nr} with LLaMA-3.1-8B (experiment details in Sec. \ref{sec:effect}). Results of ``Ours'' are aggregated from Figure \ref{fig:ablate} and \ref{fig:ablate_more}  (experiment details in Sec. \ref{sec:final_method}). The results of ``Disable Params.'' is taken from Table \ref{tab:param} with details in Sec. \ref{sec:effect_parameter}. The figure of precision under various corruption level is obtained from Figure \ref{fig:precision_50_qwen_ppl_val_ppl} using LLaMA-3.1-8B at a recall of 0.5 (check Sec. \ref{sec:filter} for more details).

\section{Dataset Details}
\subsection{Corrupted Datasets}
\label{app:data_corrupt}
\begin{figure}[t!]
\centering
\includegraphics[width=\linewidth]{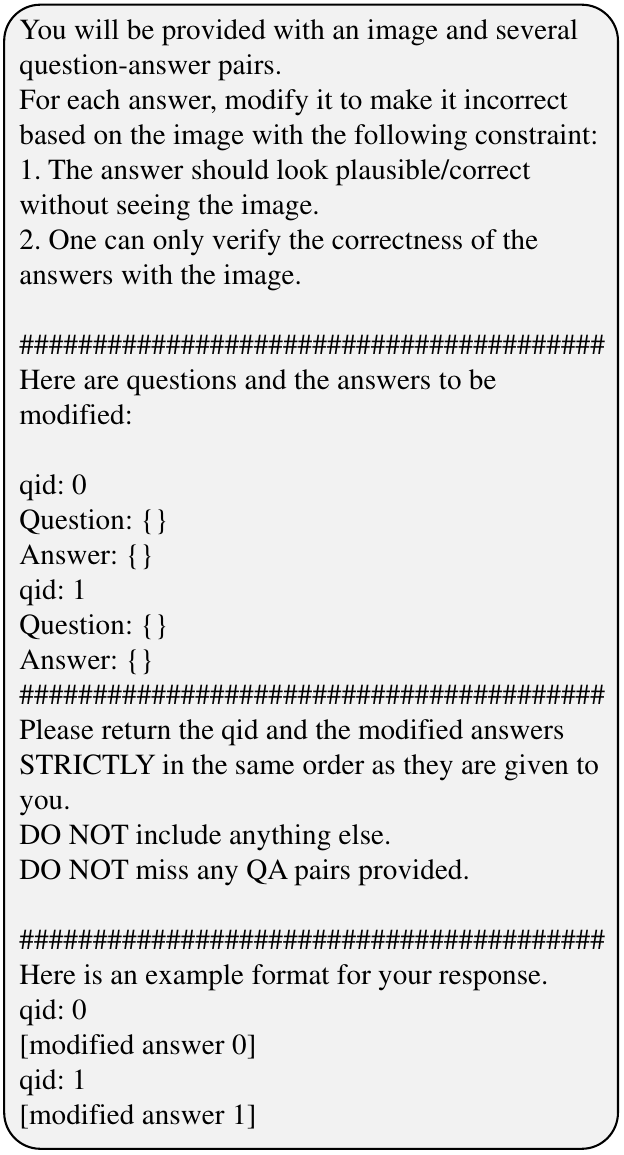}
\vspace{-.3in}
\caption{\textbf{Prompts for generating corrupted data.}}
\label{fig:prompt}
\end{figure}

\begin{figure*}[!]
\centering
\includegraphics[width=\linewidth]{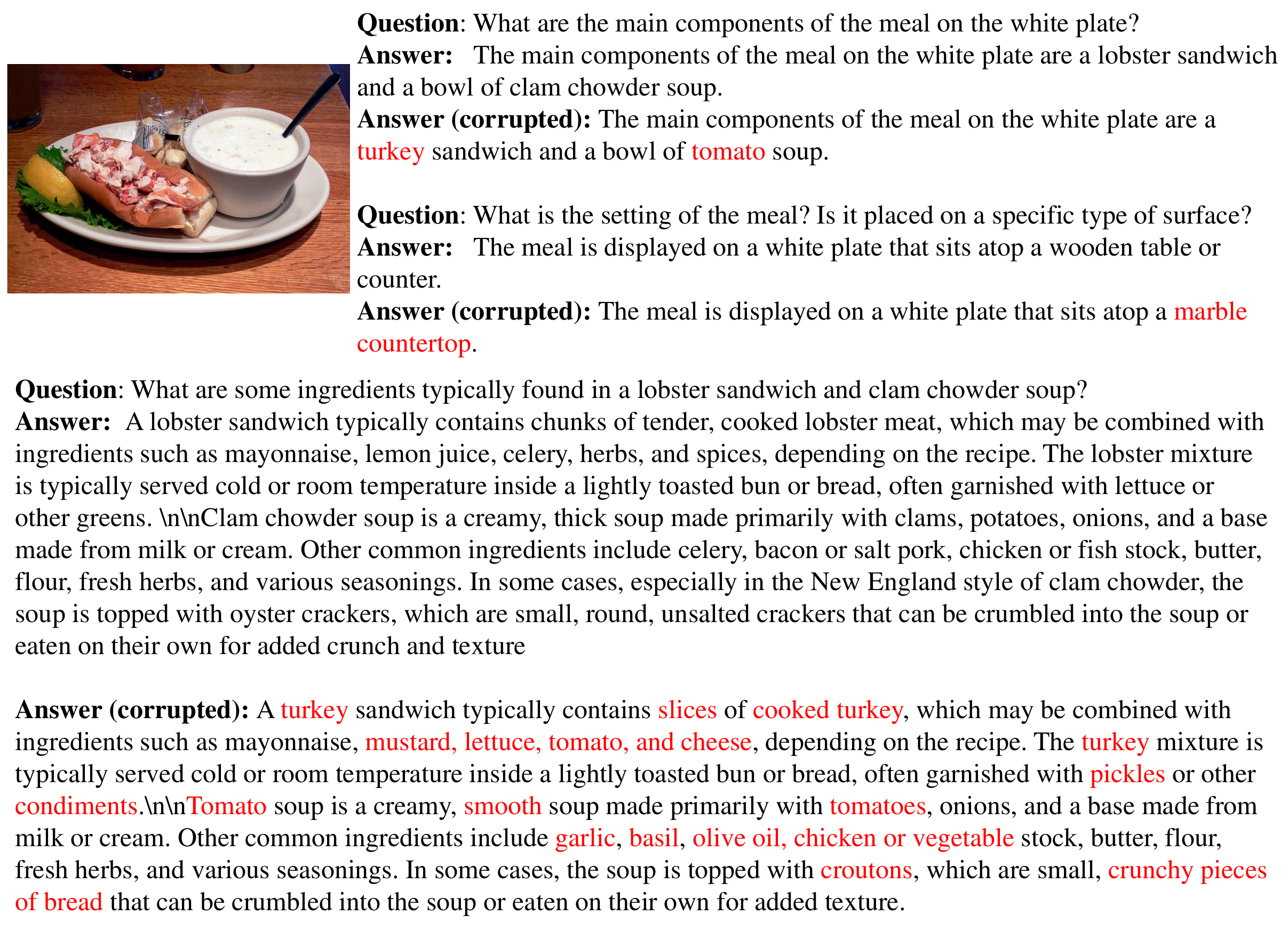}
\vspace{-.3in}
\caption{Example of corrupted sample in dataset (\textbf{LLaVA-158K}).}
\label{fig:exp_llava150}
\end{figure*}

\begin{figure*}[!]
\centering
\includegraphics[width=\linewidth]{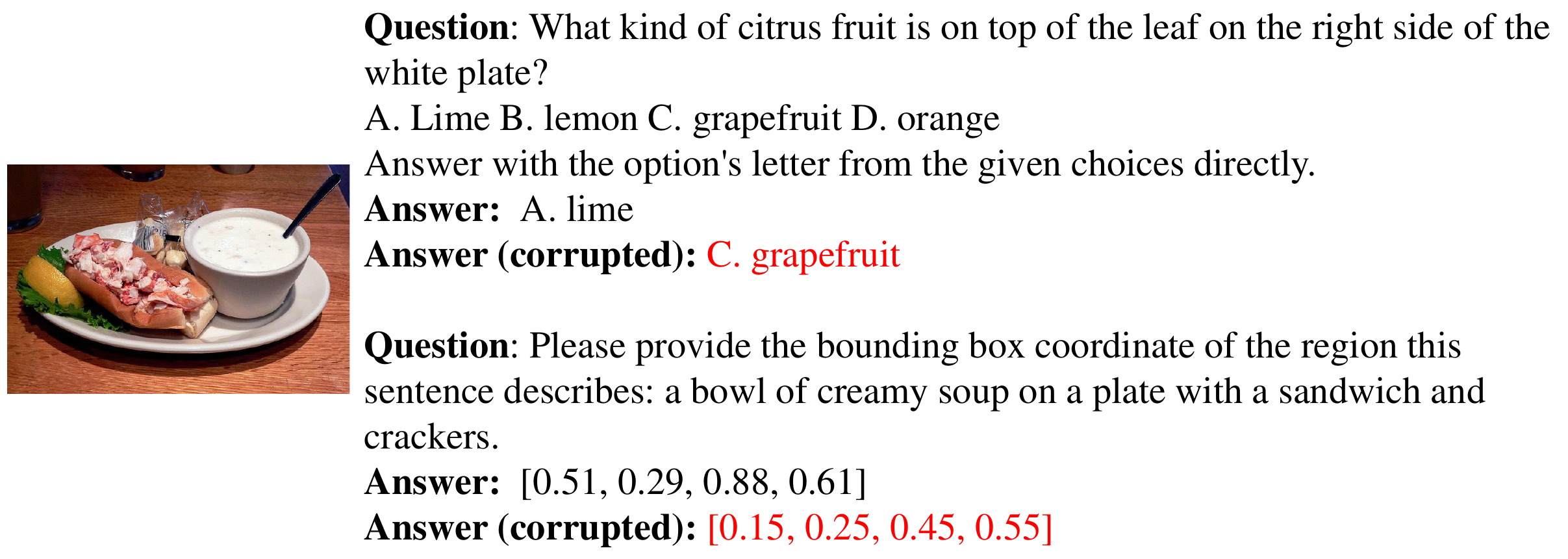}
\vspace{-.3in}
\caption{Example of corrupted sample in datasets \textbf{A-OKVQA} and \textbf{RefCOCO}. 
}
\label{fig:exp_aokvqa}
\end{figure*}

\begin{figure*}[!]
\centering
\includegraphics[width=\linewidth]{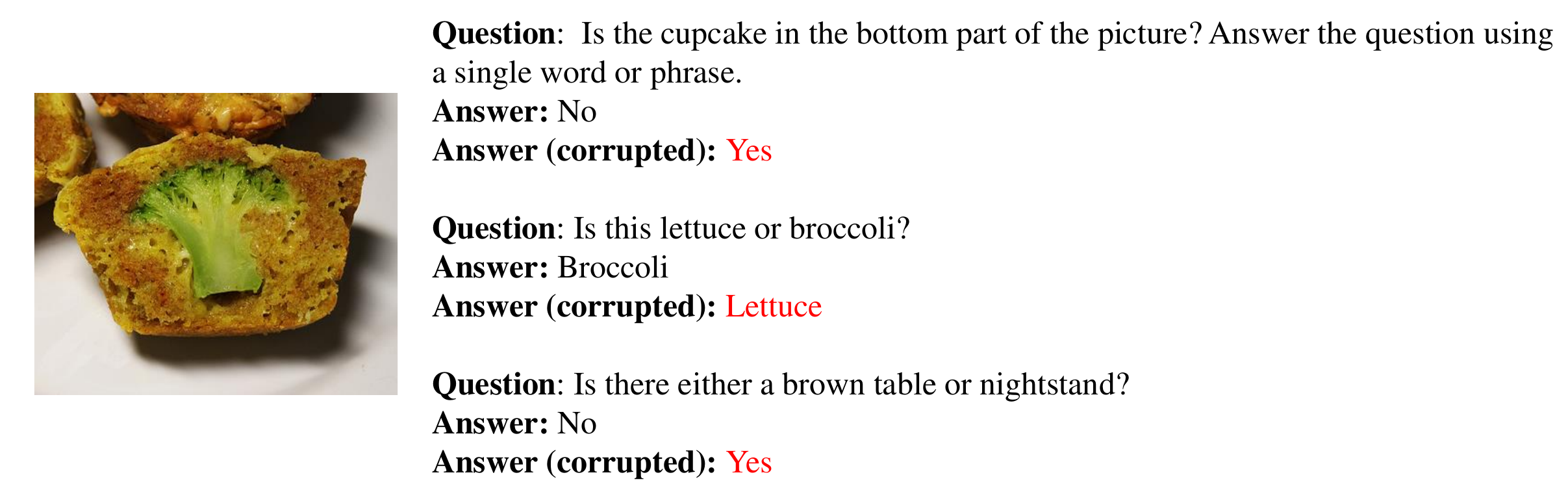}
\vspace{-.3in}
\caption{Example of corrupted sample in dataset \textbf{GQA}.}
\label{fig:exp_gqa}
\end{figure*}

\begin{figure*}[!]
\centering
\includegraphics[width=\linewidth]{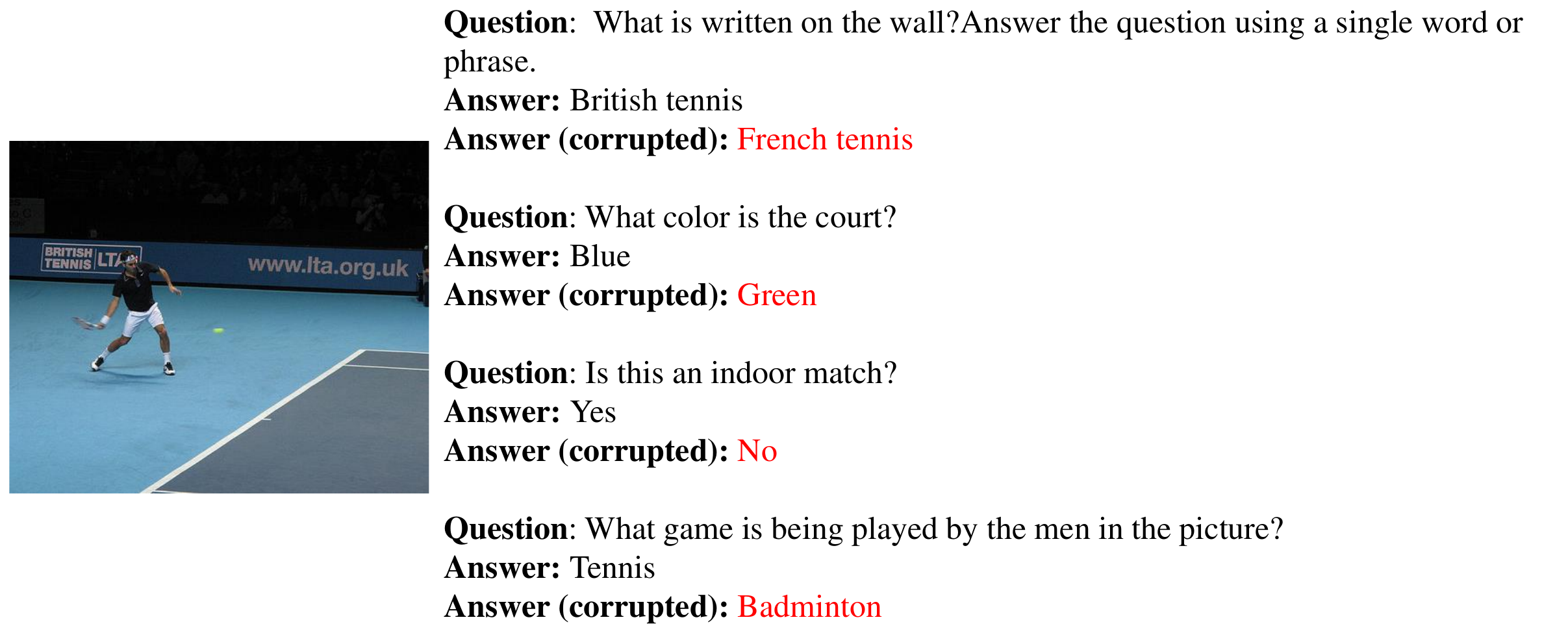}
\vspace{-.3in}
\caption{Example of corrupted sample in dataset \textbf{VQAv2}.}
\label{fig:exp_vqav2}
\end{figure*}

\begin{figure*}[!]
\centering
\includegraphics[width=\linewidth]{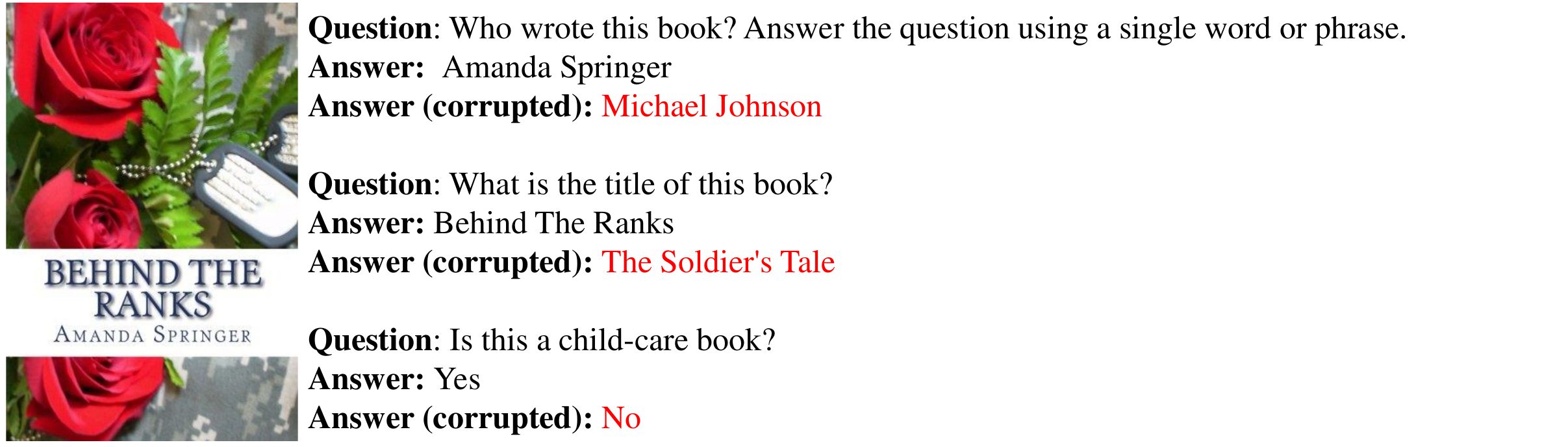}
\vspace{-.3in}
\caption{Example of corrupted sample in dataset \textbf{OCRVQA}.}
\label{fig:exp_ocrvqa}
\end{figure*}

\begin{figure*}[!]
\centering
\includegraphics[width=\linewidth]{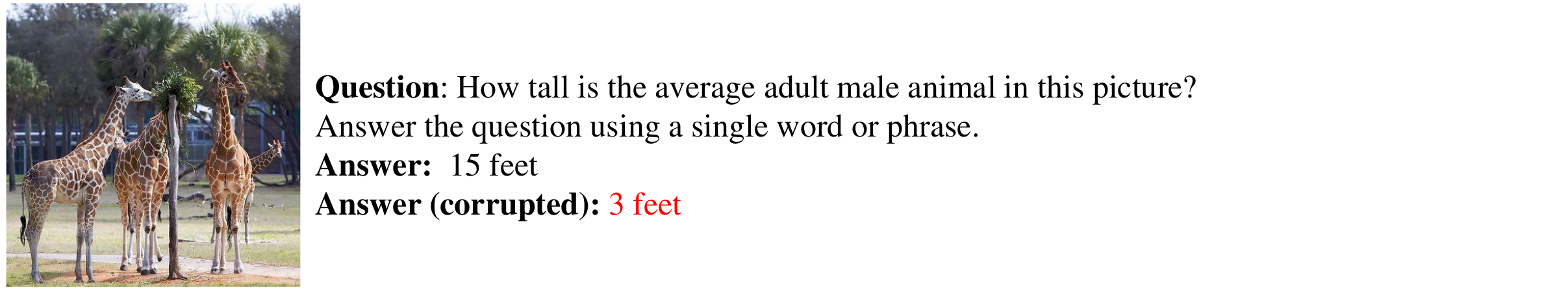}
\vspace{-.3in}
\caption{Example of corrupted sample in dataset \textbf{OKVQA}.}
\label{fig:exp_okvqa}
\end{figure*}

\begin{figure*}[!]
\centering
\includegraphics[width=\linewidth]{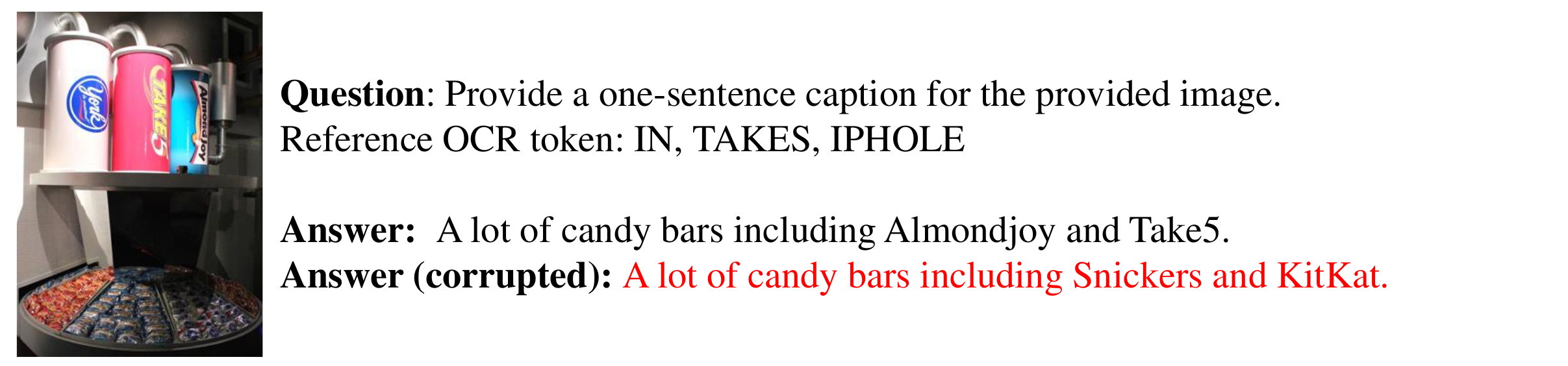}
\vspace{-.3in}
\caption{Example of corrupted sample in dataset \textbf{TextCaps}.}
\label{fig:exp_textcaps}
\end{figure*}

The prompt used by GPT-4o (described in Sec \ref{sec:preliminary}) to generate corrupted samples is shown in Figure \ref{fig:prompt}. We provide examples of corrupted samples
for each of the datasets used in Figures
\ref{fig:exp_llava150}-\ref{fig:exp_textcaps}. As can be seen, GPT produces corrupted samples in the following ways: changing the option letter in multiple choice VQA (Figure \ref{fig:exp_aokvqa}), replacing the correct answer in VQA with a plausible but incorrect one (Figures \ref{fig:exp_gqa}-\ref{fig:exp_okvqa}), inducing object hallucination in a conversation (Figure \ref{fig:exp_llava150}) and generating wrong captions (Figure \ref{fig:exp_textcaps}).



\subsection{Task Taxonomy}
\label{app:data_taxonomy}
To better understand the effect of corruption on the fine-tuning process of MLLMs, we categorize the training and evaluation tasks into 3
classes according to their response formatting prompts. These prompts specify how the MLLM should format a response when queried with a question.
\begin{itemize}
    \item VQA (visual question answering): The MLLM is prompted to answer shortly via \emph{``Answer the question using a single word or phrase.''}
    \item MC-VQA (multiple-choice VQA): The MLLM is prompted to answer only the option via \emph{``Answer with the option's letter from the given choices directly.''}
    \item Conversation: The MLLM receives no format prompt. It responds to the question in a verbose way like a conversation. 
    \item Others (not categorized): The format prompt of this task falls into none of VQA, MC-VQA and Conversation. These format prompts only appear in the training dataset.
\end{itemize}
Based on the above categories, we list the training and evaluation datasets along with our taxonomy in Tables \ref{tab:training_taxonomy} and \ref{tab:evaluation_taxonomy}.

\begin{table}[t]
\centering
\setlength{\tabcolsep}{8pt} 
\renewcommand{\arraystretch}{1.3} 
\begin{tabular}{@{}p{2cm}p{5cm}@{}}
\toprule
\textbf{Category} & \textbf{Training Datasets} \\
\midrule
\textbf{VQA}       & VQAv2~\citep{goyal2017making}, GQA~\citep{hudson2019gqa}, OKVQA~\citep{marino2019ok}, OCRVQA~\citep{8978122}       \\
\textbf{MC-VQA}    & A-OKVQA~\citep{schwenk2022okvqa}                   \\
\textbf{Conversation} & LLaVA-158K~\citep{liu2023llava}, ShareGPT~\citep{sharegpt}                         \\
\textbf{Others (not categorized)}    & TextCaps~\citep{sidorov2020textcaps}, VG~\citep{krishna2017visual}, RefCOCO~\citep{kazemzadeh-etal-2014-referitgame,mao2016generation}     \\
\bottomrule
\end{tabular}
\caption{\textbf{Taxonomy of 10 Training Datasets.} Note that no corruption is injected into ShareGPT in all our experiments as it is a text-only dataset.}
\label{tab:training_taxonomy}
\end{table}

\begin{table}[t]
\centering
\setlength{\tabcolsep}{8pt} 
\renewcommand{\arraystretch}{1.3} 
\label{tab:evaluation_taxonomy}
\begin{tabular}{@{}p{2cm}p{5cm}@{}}
\toprule
\textbf{Category} & \textbf{Evaluation Datasets} \\
\midrule
\textbf{VQA} & GQA~\citep{hudson2019gqa},  MME~\citep{fu2024mmecomprehensiveevaluationbenchmark}, POPE~\citep{li2023evaluating}, OKVQA~\citep{marino2019ok},  TextVQA~\citep{singh2019textvqa} \\ 
\textbf{MC-VQA} & MMB~\citep{liu2023mmbench}, SEED-IMG~\citep{li2023seed}, SciQA-IMG~\citep{lu2022scienceqa}    \\
\textbf{Conversation} & LLaVA-Wild~\citep{liu2023llava}, MM-Vet~\citep{yu2023mmvet} \\
\bottomrule
\end{tabular}
\caption{\textbf{Taxonomy of the 11 Evaluation Datasets.} Note that MME can be split into the perception set MME\_P and the cognition set MME\_C. }
\label{tab:evaluation_taxonomy}
\end{table}

\subsection{Evaluation Metrics}
\label{app:data_metrics}
The score ranges for MME\_P and MME\_C are $[0, 2000]$ and $[0, 800]$, respectively. In our paper, we report both their original values and the normalized values (scaled to $[0, 100]$) interchangeably. All the remaining datasets have a metric range of $[0, 100]$. We also report the average performance by taking the mean scores (normalized) of the 11 tasks in some experiments.

\section{More on Effects of Corrupted Data}
\label{app:more_effect}
\begin{figure*}[!]
\centering
\includegraphics[width=\linewidth]{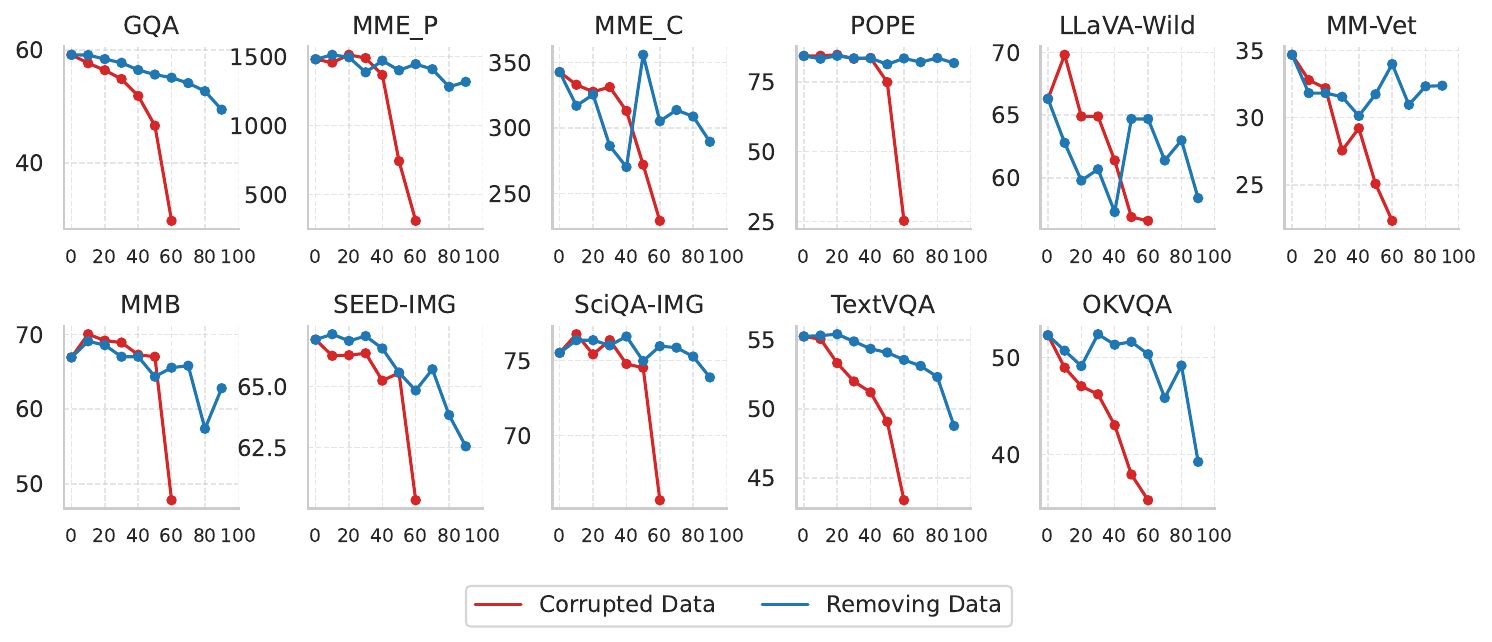}
\caption{\textbf{Performance (y-axis) of LLaVA-1.5 (LLaMA-3.1-8B) under different corruption ratios (x-axis).}}
\label{fig:effect_nr}
\end{figure*}

Due to limited space, Figure \ref{fig:effect_nr_main} only presents a subset of evaluation tasks used in this paper. Here we provide the results for all tasks in Figure \ref{fig:effect_nr}.

\section{Details on Identifying Important and Corruption-related Weights}
\label{app:prune}

\subsection{Identifying Important Weights}
\label{app:prune_snip}
Following~\cite{wei2024assessing}, we use the SNIP score~\citep{lee2018snip} to quantify weight importance. For any linear layer with a weight matrix \(W \in \mathbb{R}^{d_{\mathrm{out}} \times d_{\mathrm{in}}}\), the importance score of a weight entry \(W_{ij}\) is given by:
\[
I(W_{ij}, x) = |W_{ij} \cdot \nabla_{W_{ij}} \ell(x;\bm{\theta})|,
\]
which is a first-order Taylor approximation of the change in the loss when \(W_{ij}\) is set to zero. In matrix form, this extends to:
\[
I(W, x) = |W \odot \nabla_{W} \ell(x;\bm{\theta})|,
\]
where \(\odot\) denotes element-wise multiplication. Given a dataset \(D^{*}\) of interest, we aggregate importance scores over all instances:
\[
I(W) =  \mathbb{E}_{x \sim D^{*}} |W \odot \nabla_{W} \ell(x;\bm{\theta})|.
\]
Intuitively, \(I(W)\) measures how critical each weight is for the model's predictions on $D^{*}$. A small \(I(W)_{ij}\) indicates that setting \(W_{ij}\) to zero has minimal impact on the loss.

\subsection{Identifying Corruption-related Weights}
A straightforward approach to identifying corruption-related weights is to compute \(I(W)\) using a corrupted dataset and select the highest-ranked weights. However, some of these weights may also contribute to correct predictions. To address this, we adopt the approach proposed in \citet{wei2024assessing} to identify weights that are \textbf{specific} to corrupted data by leveraging set difference.

Specifically, we compute \(I^c\) using a model trained only with clean data and a small dataset consisting of 1K clean samples as $D^{*}$. Similarly, we compute \(I^n\) using a model trained on dataset with 60\% corruption and an 1K corrupted dataset as $D^{*}$. For any pair of sparsity levels \((p, q)\), we define the top-\(p\%\) important weights \(S^c(p)\) for \textbf{clean} samples as the weights whose \(I^c_{i,j}\) scores rank within the top \(p\%\) of the \(i\)-th row of \(I^c\)~\citep{sun2024a}:
\[
S^c(p) = \{(i,j) | I^c_{i,j} \text{ is among the top } p\% \text{ of } I^c_i\}.
\]
Similarly, we define the top-\(q\%\) important weights \(S^n(q)\) for \textbf{corrupted} samples as :
\[
S^n(q) = \{(i,j) | I^n_{i,j} \text{ is among the top } q\% \text{ of } I^n_i\}.
\]
Finally, the isolated weights \(S(p,q)\) are defined as the set difference between \(S^n(q)\) and \(S^c(p)\):
\[
S(p,q) = S^n(q) - S^c(p).
\]
This approach isolates weights specific to corrupted samples while filtering out those that are also important for producing clean samples.

\paragraph{Choices of $(p,q)$.}
We begin with small, identical values for $(p,q)$ and gradually increase them until we observe a significant performance improvement compared to the original model. To prevent the model from collapsing due to the removal of critical weights essential for correct predictions, we set $p$ to be slightly larger than $q$. This ensures that more of the weights responsible for generating correct, clean responses are preserved. We find that this approach leads to better overall performance, with fewer parameters disabled.

\section{Details on Identifying Correct Samples using MLLM}
\subsection{Details on Scores}
\label{app:ppl}
\paragraph{Perplexity}
For a sample $x$, its perplexity is computed as : 
\begin{equation}
    \texttt{PPL}(x) = \exp\left(-\frac{1}{n} \sum_{t=1}^{n} \log p(x_y^t | x_y^{<t}, x_c; \bm{\theta})\right),
\end{equation}
where $n$ is the length of the response.
It quantifies how ``surprised'' or ``uncertain'' a model is when generating a response conditioned on an instruction. A lower PPL indicates higher confidence, while a higher PPL suggests that the response is less probable under the model’s learned distribution. Intuitively, a higher PPL indicates the sample is more likely to be corrupted according to the models' knowledge.  

The loss for this sample is $\log{\texttt{PPL}(x)}$. 

\subsection{Decision Rule and Evaluation}
\label{app:classify}
Giving a score (\eg, \texttt{PPL} and \texttt{Val\_PPL}) and the dataset (100K samples with $cr=50\%$), the following is conducted: 
\begin{enumerate}
    \item Computing the score for all the samples in a dataset.
    \item Choosing thresholds \( \tau \) starting from the 1st to the 100th lower percentiles of the score. Then, for each threshold \( \tau \), we define: $\hat{z}_i = \mathbbm{1} \left( \text{score}(x_i) < \tau \right),$ where \( \mathbbm{1}(\cdot) \) is the indicator function. Intuitively, a sample $x_i$ is predicted as clean $\hat{z}_i = 1$ if its score is lower than the threshold. 
    \item For $N$ samples in total, we compute precision ($P$) and recall ($R$) scores for all thresholds as follows:
\[
P = \frac{ \sum_{i=1}^N \mathbbm{1}(z_i = 1 \text{ and } \hat{z}_i = 1) }{ \sum_{i=1}^N \mathbbm{1}( \hat{z}_i = 1 ) }.
\]

$$
R = \frac{ \sum_{i=1}^N \mathbbm{1}(z_i = 1 \text{ and } \hat{z}_i = 1) }{ \sum_{i=1}^N \mathbbm{1}(z_i = 1) }.
$$
    \item We draw the precision-recall curve.
\end{enumerate}

\subsection{Analysis on Improved Understanding of Corrupted Samples}
\label{app:analysis_improved}
To understand why improved understanding of corrupted samples lead to the distribution shift of \( \hat{p}(z=1|x_c, \tilde{x}_y; \bm{\theta}) \) on corrupted samples, we start by rewriting $p(z = 1 \mid x_c, \tilde{x}_y)$
using Bayes' rule as
\begin{align} 
    p(z = 1 \mid x_c, \tilde{x}_y) =  1- p(z = 0 \mid x_c, \tilde{x}_y).
    \label{eq:bayes}
\end{align}

By Bayes' theorem, we have
\begin{align}
    & p(z = 0 \mid x_c, \tilde{x}_y) \notag \\
     = & \frac{p(\tilde{x}_y \mid z = 0, x_c)}{p(\tilde{x}_y \mid x_c)} \cdot p(z = 0 \mid x_c).
    \label{eq:tmp}
\end{align}

Since corruption is uniformly applied to all samples, the correctness label \( z \) is independent of \( x_c \), and thus
\[
p(z = 0 \mid x_c) = p(z = 0).
\]

Substituting this into (\ref{eq:bayes}) and (\ref{eq:tmp}), we get
\[
p(z = 1 \mid x_c, \tilde{x}_y) = 1 - c \cdot \frac{p(\tilde{x}_y \mid x_c, z = 0)}{p(\tilde{x}_y \mid x_c)},
\]
where \( c = p(z = 0) \).
Therefore,
\[
p(z = 1 \mid x_c, \tilde{x}_y) - 1 \propto - \frac{p(\tilde{x}_y \mid x_c, z = 0)}{p(\tilde{x}_y \mid x_c)}.
\]

This shows that 
\( p(z = 1 | x_c, \tilde{x}_y) -1\) is proportional
to the negative of the ratio between 
\( p(\tilde{x}_y | x_c, z = 0) \) 
and 
\( p(\tilde{x}_y | x_c) \).  
Note \( p(\tilde{x}_y | x_c, z = 0) \) can be regarded as the MLLM's understanding of corrupted samples, which 
we estimate through explicitly prompting the MLLM similar to \texttt{Val\_PPL} as follows:

\begin{promptcond*}{}{}
{\textit{<image>Give me an \textcolor{red}{incorrect} answer for the following question.} \\$\left\{\text{instruction text}\right\}$}
\end{promptcond*}  

\begin{figure}[t!]
\centering
\includegraphics[width=\linewidth]{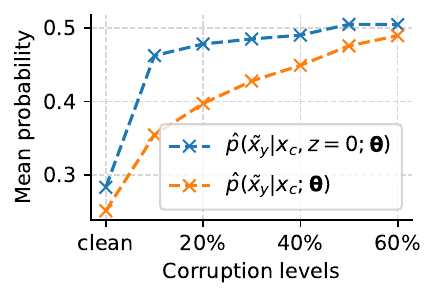}
\caption{Mean of $\hat{p}(\tilde{x}_y | x_c, z=0; \bm{\theta})$ and $\hat{p}(\tilde{x}_y | x_c; \bm{\theta})$ of corrupted samples.}
\label{fig:cond_likelihood_clean}
\end{figure}

Figure~\ref{fig:cond_likelihood_clean}
shows the means of the estimated probabilities
$\hat{p}(\tilde{x}_y | x_c, z=0; \bm{\theta})$ and $\hat{p}(\tilde{x}_y | x_c; \bm{\theta})$ at \mbox{varying} corruption levels over the corrupted datasets for fine-tuning.
We observe a sharp rise in $\hat{p}(\tilde{x}_y | x_c, z=0; \bm{\theta})$ (\textcolor{color1}{blue}) when transitioning from a clean model to one trained with slight corruption (\( cr=10\% \)), followed by a slower growth as the corruption increases further. In contrast, $\hat{p}(\tilde{x}_y | x_c; \bm{\theta})$ (\textcolor{color2}{orange}) exhibits a modest increase and remains consistently lower than the other probability.

This shows the MLLM's understanding of corrupted $\hat{p}(\tilde{x}_y | x_c, z=0; \bm{\theta})$ gets significantly improved when only including 10\% of corruptions. 

\section{Implementation Details on Baselines}
\label{app:baseline}
\subsection{Noise-robust Loss Functions}
\label{app:loss_func}
\begin{figure}
\centering
\includegraphics[width=0.5\textwidth]{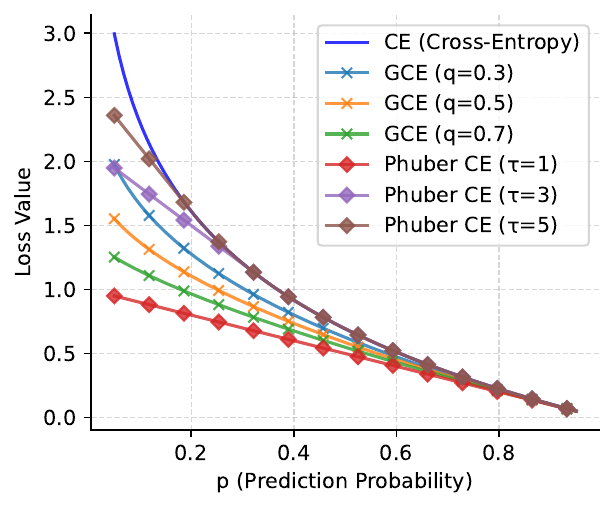} 
\caption{\textbf{Visualization of noise-robust loss functions.}}
\label{fig:robust_loss}
\end{figure}

The mathematical formulations for the loss functions and their corresponding hyper-parameters are provided in Table \ref{tab:loss}. For GCE, a larger $q$ ($q \in (0, 1]$) reduces the sensitivity of the loss to over-confident yet incorrect predictions. In the case of Phuber CE, the hyper-parameter $\tau$ ($\tau > 1$) determines the threshold for gradient clipping when the model produces over-confident but incorrect predictions, thereby enhancing its robustness to potential data noise. Overall, these loss functions aim to mitigate the adverse impact of small predicted probabilities for the target classes (see loss visualizations in Figure \ref{fig:robust_loss}), which often arise due to corrupted data. To optimize performance, we conduct a hyper-parameter search for $q$ and $\tau$ in Figure \ref{fig:hyper}.

\subsection{Sample Selection Methods}
\label{app:sample_select}
\begin{figure*}[!]
\centering
\includegraphics[width=\linewidth]{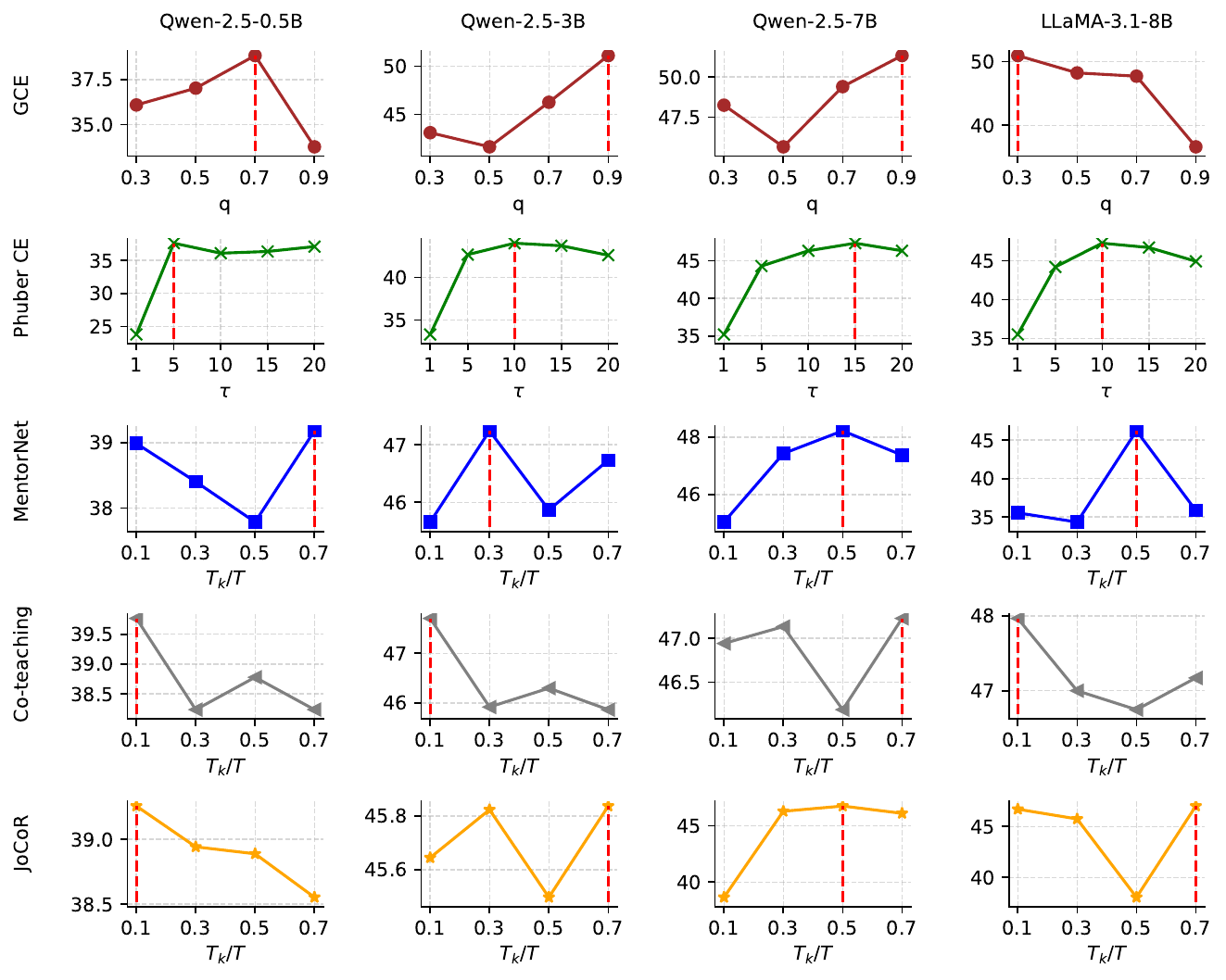}
\vspace{-.3in}
\caption{\textbf{Ablations on the choices of hyper-parameters for noise-robust loss functions and sample selection methods.} Average performance are reported. All experiments are conducted on datasets with $cr=50\%$ and best results are indicated by red dashed lines and reported in Tables \ref{tab:main_llama3}, \ref{tab:main_qwen_0.5b}, \ref{tab:main_qwen_3b} and \ref{tab:main_qwen_7b}}
\label{fig:hyper}
\end{figure*}

For MentorNet, Co-teaching and JoCoR (Algorithms \ref{alg:mentornet}, \ref{alg:coteaching} and \ref{alg:jocor}). We assume access to an estimated noise level $\alpha$ for the dataset and drop certain amount of data (which is linearly warmed up from 0 to $\alpha$ in $T_k$ steps) according to the defined criterion in the algorithm. Recall the definition of loss in Sec. \ref{sec:preliminary}, we let $\gL(\mathcal{B}; \bm{\theta})$ be the batch-wise formulation $\sum_{x\in \mathcal{B}}\frac{1}{|\mathcal{B}|}\ell(x;\bm{\theta})$. We search for the best $\frac{T_k}{T}$ in Figure \ref{fig:hyper}.

\paragraph{MentorNet} While various implementations of MentorNet exist, we follow prior works on learning with label noise~\cite{han2018co, wei2020combating} and adopt a simple selection criterion based on self-computed loss. Specifically, we retain only samples with small loss for model training.

\paragraph{Co-teaching} Originally, Co-teaching utilizes two randomly initialized networks to generate diverse predictions and mitigate error accumulation. However, in the era of MLLM, pre-trained LLMs are required for initialization. To introduce diversity in model predictions, we train them using two independently shuffled data-loaders (with different random seeds). This strategy is also employed in JoCoR, which similarly relies on two distinct networks.

\paragraph{JoCoR}
It was originally designed for image classification tasks, where the consistency loss minimizes the divergence between two models' class predictions $y$ for images $x$:
\[
\ell_{con}^{\text{cls}}(x; \bm{\theta}_f, \bm{\theta}_g) = D_{KL}(p_f \| p_g) + D_{KL}(p_g \| p_f), 
\]
\[
p_f = p(y | x; \bm{\theta}_f), \quad p_g = p(y | x; \bm{\theta}_g).
\]
For autoregressive sequence generation, the model predicts a sequence token by token, requiring consistency at each step:
\[
\ell_{con}^{\text{seq}}(x; \bm{\theta}_f) = \sum_{t=1}^{T} D_{KL}(p_f^t \| p_g^t) + D_{KL}(p_g^t \| p_f^t).
\]
where
\[
p_f^t = p(x_y^t | x_y^{<t}, x_c; \bm{\theta}_f), 
\]
\[
p_g^t = p(x_y^t | x_y^{<t}, x_c; \bm{\theta}_g).
\]
This ensures divergence minimization at each decoding step rather than a single output. Therefore, we use $\ell_{con}^{\text{cls}}$ as $\ell_{con}$ and $\gL_{con}(\mathcal{B}; \bm{\theta}_f, \bm{\theta}_g)$ its batch-wise formulation.

\subsection{Scores Used for Sample Selection in Further Fine-tuning}
\label{app:sample_select_further}
Let \( p_t(i) = p(x_y^t = i | x_y^{<t}, x_c; \bm{\theta}) \) denote the model's predicted probability for token \( i \) at timestep \( t \). 

\paragraph{Entropy}
The entropy is computed as   
\[
H(x; \bm{\theta}) = - \frac{1}{T} \sum_{t=1}^{T} \sum_{i=1}^{V} p_t(i) \log p_t(i),
\]
which measures the model’s uncertainty, averaged over tokens.

\paragraph{EL2N}
Denoting one-hot indicator vector for the true token \( x_y^t \) as  
\[
\mathbf{1}_{x_y^t} \in \mathbb{R}^{V}, \quad \text{where} \quad \mathbf{1}_{x_y^t, i} =
\begin{cases}
1, & \text{if } x_y^t = i, \\
0, & \text{otherwise}.
\end{cases}
\]
where \( V \) is the vocabulary size. 
The  L2-norm of the output error (EL2N) is computed as:  
\[
\text{EL2N}(x; \bm{\theta}) = \frac{1}{T} \sum_{t=1}^{T} \sqrt{\sum_{i=1}^{V} \left( p_t(i) - \mathbf{1}_{x_y^t, i} \right)^2}.
\]
which quantifies the deviation of the predicted probability distribution from the ground truth, averaged over tokens.

\paragraph{GradNorm} It is defined as the L2-norm of the gradient vector that is formed by stacking the flattened gradient of each trainable parameter in the model. 
\[
\text{GradNorm}(x; \bm{\theta}) = ||\nabla_{\bm{\theta}}\ell(x_y| x_c; \bm{\theta})||_2
\]

When using these scores for sample selection in further fine-tuning, we experimented with both higher and lower values and found that selecting samples with lower values yielded better results (Table \ref{tab:ablation_lower_higher}).

\begin{table*}[h!]
\centering
\begin{tabular}{lccc}
\toprule
Loss    & Definition  & Hyper-parameters \\ 
\midrule
CE & $-\log(p)$ & - \\
GCE   & $(1-p^q)/q$   & $q \in (0,1]$           \\ 
Phuber CE    & $\begin{cases}
    -\log(p) & p > \frac{1}{\tau} \\
    1 + \log(\tau) - \tau*p  & p \le \frac{1}{\tau}
\end{cases} $     & $\tau > 1$ \\ 
                            \bottomrule
\end{tabular}%
\caption{Mathematical definition for noise robust loss functions. $p$ denotes the probability of predicting specific tokens.}
\label{tab:loss}
\end{table*}

\begin{algorithm*}[ht]
	\caption{MentorNet}
	\begin{algorithmic}[1]
		\STATE Let $\bm{\theta}$ be the MLLM parameters, $T$ the total number of training steps, $\mathcal{D}$ the corrupted dataset, $\alpha$ the estimated corruption level, $\eta$ the learning rate, and $T_k$ the warm-up steps;
            \STATE Shuffle training data $\mathcal{D}$;
		\FOR{$t = 0, \dots, T-1$}
            \STATE $R(t) = 1 - \min \left\{\frac{t}{T_k} \cdot \alpha, \alpha\right\}$;
		\STATE Draw a mini-batch $\mathcal{B}$ from the dataset $\mathcal{D}$;
            \STATE Select $R(t) \cdot |\mathcal{B}|$ small-loss samples $\hat{\mathcal{B}}$ from $\mathcal{B}$ based on $\ell(x_i; \bm{\theta})$;
		\STATE Update the parameters: $\bm{\theta} = \bm{\theta} - \eta \nabla_{\bm{\theta}} \gL(\hat{\mathcal{B}}; \bm{\theta})$;
		\ENDFOR
	\end{algorithmic}
	\label{alg:mentornet}
\end{algorithm*}


\begin{algorithm*}[ht]
	\caption{Co-teaching}
	\begin{algorithmic}[1]
		\STATE Let $\bm{\theta}_f$ and $\bm{\theta}_g$ be the parameters for two identical MLLMs,  $T$ the total number of training steps, $\mathcal{D}$ the corrupted dataset, $\alpha$ the estimated corruption level, $\eta$ the learning rate, and $T_k$ the warm-up steps;
		\FOR{$t = 0, \dots, T-1$}
            \STATE Shuffle training data $\mathcal{D}$ twice with different seeds to get $\mathcal{D}_f$ and $\mathcal{D}_g$;
            \STATE $R(t) = 1 - \min \left\{\frac{t}{T_k} \cdot \alpha, \alpha\right\}$;
		\STATE Draw mini-batches $\mathcal{B}_f$ and $\mathcal{B}_g$ from datasets $\mathcal{D}_f$ and $\mathcal{D}_g$;
            \STATE Select $R(t) \cdot |\mathcal{B}_g|$ small-loss samples $\hat{\mathcal{B}}_g$ from $\mathcal{B}_g$ based on $\ell(x_i; \bm{\theta}_f)$;
            \STATE Select $R(t) \cdot |\mathcal{B}_f|$ small-loss samples $\hat{\mathcal{B}}_f$ from $\mathcal{B}_f$ based on $\ell(x_i; \bm{\theta}_g)$;
		\STATE Update the parameters: $\bm{\theta}_f = \bm{\theta}_f - \eta \nabla_{\bm{\theta}_f} \gL(\hat{\mathcal{B}_f}; \bm{\theta}_f)$;
            \STATE Update the parameters: $\bm{\theta}_g = \bm{\theta}_g - \eta \nabla_{\bm{\theta}_g} \gL(\hat{\mathcal{B}_g}; \bm{\theta}_f)$;
		\ENDFOR
	\end{algorithmic}
	\label{alg:coteaching}
\end{algorithm*}


\begin{algorithm*}[ht]
	\caption{JoCoR}
	\begin{algorithmic}[1]
		\STATE Let $\bm{\theta}_f$ and $\bm{\theta}_g$ be the parameters for two identical MLLMs,  $T$ the total number of training steps, $\mathcal{D}$ the corrupted dataset, $\alpha$ the estimated corruption level, $\eta$ the learning rate, $\lambda$ the weighting co-efficient, and $T_k$ the warm-up steps;
		\FOR{$t = 0, \dots, T-1$}
            \STATE Shuffle training data $\mathcal{D}$ twice with different seeds to get $\mathcal{D}_f$ and $\mathcal{D}_g$;
            \STATE $R(t) = 1 - \min \left\{\frac{t}{T_k} \cdot \alpha, \alpha\right\}$;
		\STATE Draw mini-batches $\mathcal{B}_f$ and $\mathcal{B}_g$ from datasets $\mathcal{D}_f$ and $\mathcal{D}_g$;
            \STATE Select $R(t) \cdot |\mathcal{B}_g|$ small-loss samples $\hat{\mathcal{B}}_g$ from $\mathcal{B}_g$ based on $(1-\lambda)\ell(x_i; \bm{\theta}_f) + \lambda \ell_{con}(x_i; \bm{\theta}_f, \bm{\theta}_g)$;
            \STATE Select $R(t) \cdot |\mathcal{B}_f|$ small-loss samples $\hat{\mathcal{B}}_f$ from $\mathcal{B}_f$ based on $(1-\lambda)\ell(x_i; \bm{\theta}_g) + \lambda\ell_{con}(x_i; \bm{\theta}_g, \bm{\theta}_f)$;
		\STATE Update the parameters: $\bm{\theta}_f = \bm{\theta}_f - \eta \nabla_{\bm{\theta}_f} \left((1-\lambda)\gL(\hat{\mathcal{B}_f}; \bm{\theta}_f) + \lambda\gL_{con}(\hat{\mathcal{B}_f}; \bm{\theta}_g, \bm{\theta}_f)\right)$;
            \STATE Update the parameters: $\bm{\theta}_g = \bm{\theta}_g - \eta \nabla_{\bm{\theta}_g} \left((1-\lambda)\gL(\hat{\mathcal{B}_g}; \bm{\theta}_g) + \lambda\gL_{con}(\hat{\mathcal{B}_g}; \bm{\theta}_f, \bm{\theta}_g)\right)$;
		\ENDFOR
	\end{algorithmic}
	\label{alg:jocor}
\end{algorithm*}


\begin{table*}[h!]
\centering
\resizebox{\textwidth}{!}{%
\begin{tabular}{lcccccccccccc}
\toprule
Methods & \textbf{Avg.} & GQA & MME\_P & MME\_C & POPE & \setlength\extrarowheight{0pt}\begin{tabular}[c]{@{}c@{}}LLaVA\\ Wild\end{tabular} & MM-Vet & MMB & \setlength\extrarowheight{0pt}\begin{tabular}[c]{@{}c@{}}SEED\\ IMG\end{tabular}  & \setlength\extrarowheight{0pt}\begin{tabular}[c]{@{}c@{}}SciQA\\ IMG\end{tabular}  &  \setlength\extrarowheight{0pt}\begin{tabular}[c]{@{}c@{}}Text\\ VQA\end{tabular} & OKVQA \\ \midrule

\rowcolor{baselinecolor}
Clean & 45.78  & 47.88  & 1140.91  & 257.86  & 81.75  & 46.60  & 14.72  & 45.19  & 51.14  & 60.44  & 36.61  & 30.02   \\ 
None (CE) & 36.01  & 37.06  & 670.50  & 217.86  & 48.25  & 40.80  & 15.87  & 33.25  & 48.61  & 55.33  & 32.21  & 23.95   \\ \midrule
\multicolumn{12}{l}{\textit{Noise-robust loss functions}} \\
GCE & 38.83  & 38.23  & 735.12  & 254.29  & 46.56  & \textbf{44.80}  & \textbf{19.95}  & \textbf{40.98}  & \textbf{50.17}  & \textbf{60.59}  & \textbf{36.32}  & 20.94   \\ 
Phuber CE & 37.61  & 36.42  & 776.03  & 211.43  & 55.75  & \underline{44.20}  & 16.10  & 33.59  & 49.10  & \underline{59.44}  & 33.99  & 19.93   \\ \midrule
\multicolumn{12}{l}{\textit{Sample selection (online)}} \\
MentorNet & 39.18  & 37.54  & 882.57  & \textbf{268.57}  & 68.96  & 36.00  & 13.44  & 36.60  & 48.02  & 55.18  & 32.49  & 25.10   \\ 
Co-teaching & 39.76  & 36.85  & 884.94  & 235.71  & 69.98  & 39.20  & 18.67  & \underline{37.97}  & 48.18  & 55.13  & 32.52  & 25.20   \\ 
JoCoR & 39.26  & 36.80  & 844.99  & 239.29  & 70.40  & 38.10  & 15.37  & 36.43  & 47.61  & 56.72  & 32.95  & \underline{25.29}   \\ \midrule
\multicolumn{12}{l}{\textit{Further fine-tuning}} \\
EL2N & 36.88  & 37.19  & 646.61  & 205.36  & 61.13  & 35.00  & 16.01  & 36.86  & 48.35  & 54.73  & 33.96  & 24.40   \\ 
GradNorm & \underline{40.32}  & \underline{40.90}  & 863.06  & 233.57  & 72.08  & 43.80  & 16.42  & 36.68  & \underline{49.32}  & 57.86  & 32.80  & 21.33   \\ 
Entropy & 30.18  & 37.92  & 841.09  & 253.21  & 27.87  & 38.80  & 16.01  & 0.26  & 42.13  & 46.41  & 29.20  & 19.69   \\ 
PPL & 39.26  & 39.86  & \underline{913.69}  & 227.14  & \underline{74.46}  & 41.20  & 16.70  & 26.12  & 46.71  & 54.54  & 33.86  & 24.32   \\ 
\rowcolor{backcolor}
Val\_PPL & \textbf{42.28}  & \textbf{43.62}  & \textbf{1076.64}  & 226.79  & \textbf{77.19}  & 42.70  & \underline{18.72}  & 34.97  & 49.10  & 55.68  & \underline{35.02}  & \textbf{25.86}   \\ 
\bottomrule

\end{tabular}%
}
\caption{\textbf{Comparisons of different corruption-robust strategies at a corruption ratio of 50\% on LLaVA-1.5 (Qwen-2.5-0.5B).} Here, \textbf{Avg.} refers to the average performance on 11 benchmarks (normalized to 0-100). Best results are \textbf{Bold}, second best are \underline{underlined}.}
\label{tab:main_qwen_0.5b}
\end{table*}



\begin{table*}[h!]
\centering
\resizebox{\textwidth}{!}{%
\begin{tabular}{lcccccccccccc}
\toprule
Methods & \textbf{Avg.} & GQA & MME\_P & MME\_C & POPE & \setlength\extrarowheight{0pt}\begin{tabular}[c]{@{}c@{}}LLaVA\\ Wild\end{tabular} & MM-Vet & MMB & \setlength\extrarowheight{0pt}\begin{tabular}[c]{@{}c@{}}SEED\\ IMG\end{tabular}  & \setlength\extrarowheight{0pt}\begin{tabular}[c]{@{}c@{}}SciQA\\ IMG\end{tabular}  &  \setlength\extrarowheight{0pt}\begin{tabular}[c]{@{}c@{}}Text\\ VQA\end{tabular} & OKVQA \\ \midrule

\rowcolor{baselinecolor}
Clean & 55.84  & 53.78  & 1299.26  & 288.93  & 83.41  & 62.80  & 30.05  & 62.80  & 63.50  & 72.98  & 48.26  & 35.62   \\ 
None (CE) & 43.71  & 37.40  & 792.42  & 243.93  & 32.06  & 59.00  & 19.54  & 56.62  & 62.86  & 71.64  & 42.07  & 29.56   \\ \midrule
\multicolumn{12}{l}{\textit{Noise-robust loss functions}} \\
GCE & \underline{51.11}  & 44.26  & \underline{1203.31}  & 249.64  & 76.49  & 58.30  & \underline{28.26}  & 56.62  & 56.55  & 68.22  & 45.40  & \underline{36.79}   \\ 
Phuber CE & 44.00  & 37.25  & 795.80  & 234.64  & 21.00  & 60.00  & 24.95  & 60.22  & 63.10  & 72.04  & \underline{45.50}  & 30.77   \\ \midrule
\multicolumn{12}{l}{\textit{Sample selection (online)}} \\
MentorNet & 47.24  & 37.75  & 944.09  & 249.29  & 61.47  & 57.00  & 24.04  & 56.53  & 56.71  & \underline{73.28}  & 42.62  & 31.92   \\ 
Co-teaching & 47.70  & 37.32  & 805.62  & 252.50  & 58.10  & \underline{60.40}  & 27.71  & 59.79  & 60.60  & \textbf{73.82}  & 42.28  & 32.85   \\ 
JoCor & 45.84  & 37.46  & 658.20  & 221.43  & 57.53  & 57.50  & 24.68  & 58.59  & 61.04  & 72.04  & 42.67  & 32.10   \\ 
\midrule
\multicolumn{12}{l}{\textit{Further fine-tuning}} \\
EL2N & 51.06  & 42.49  & 1060.59  & 265.36  & 74.88  & 57.70  & 23.72  & 63.83  & \underline{63.16}  & 72.83  & 43.11  & 33.75   \\ 
GradNorm & 49.49  & 43.60  & 909.99  & \textbf{303.93}  & \underline{80.27}  & 55.00  & 24.86  & 56.44  & 61.90  & 70.90  & 42.95  & 24.96   \\ 
Entropy & 48.81  & 42.11  & 996.08  & 266.43  & 78.76  & 55.50  & 20.37  & 54.30  & 59.33  & 70.05  & 42.98  & 30.45   \\ 
PPL & 46.87  & 40.17  & 1126.94  & 231.79  & 35.05  & 57.60  & 24.86  & \textbf{65.64}  & 63.11  & 72.63  & 41.77  & 29.37   \\ 
\rowcolor{backcolor}
Val\_PPL & \textbf{55.70}  & \textbf{52.03}  & \textbf{1254.83}  & \underline{287.14}  & \textbf{82.66}  & \textbf{60.60} & 27.11  & \underline{65.29}  & \textbf{63.51}  & 72.63  & \textbf{48.21}  & \textbf{42.07}   \\ 
\bottomrule
        
\end{tabular}%
}
\caption{\textbf{Comparisons of different corruption-robust strategies at a corruption ratio of 50\% on LLaVA-1.5 (Qwen-2.5-3B).} Here, \textbf{Avg.} refers to the average performance on 11 benchmarks (normalized to 0-100). Best results are \textbf{Bold}, second best are \underline{underlined}.}
\label{tab:main_qwen_3b}
\end{table*}


\begin{table*}[h!]
\centering
\resizebox{\textwidth}{!}{%
\begin{tabular}{lcccccccccccc}
\toprule
Methods & \textbf{Avg.} & GQA & MME\_P & MME\_C & POPE & \setlength\extrarowheight{0pt}\begin{tabular}[c]{@{}c@{}}LLaVA\\ Wild\end{tabular} & MM-Vet & MMB & \setlength\extrarowheight{0pt}\begin{tabular}[c]{@{}c@{}}SEED\\ IMG\end{tabular}  & \setlength\extrarowheight{0pt}\begin{tabular}[c]{@{}c@{}}SciQA\\ IMG\end{tabular}  &  \setlength\extrarowheight{0pt}\begin{tabular}[c]{@{}c@{}}Text\\ VQA\end{tabular} & OKVQA \\ \midrule

\rowcolor{baselinecolor}
Clean & 59.75  & 56.94  & 1479.14  & 292.50  & 84.76  & 65.00  & 34.08  & 70.19  & 66.16  & 76.30  & 52.71  & 40.57   \\ 
None (CE) & 45.56  & 38.65  & 765.30  & 273.57  & 26.31  & 55.20  & 23.49  & 64.78  & \underline{65.64}  & 73.72  & 46.25  & 34.62   \\ \midrule
\multicolumn{12}{l}{\textit{Noise-robust loss functions}} \\
GCE & 51.34  & 43.66  & 1129.36  & 291.43  & 68.83  & 57.70  & 27.80  & 57.73  & 57.15  & 72.24  & \underline{49.38}  & 37.34   \\ 
Phuber CE & 47.32  & 39.73  & 848.28  & 269.64  & 31.47  & \underline{60.10}  & \underline{28.35}  & 64.78  & 64.46  & 73.57  & 49.04  & 32.93   \\  \midrule
\multicolumn{12}{l}{\textit{Sample selection (online)}} \\
MentorNet & 48.22  & 40.98  & 1055.65  & 280.36  & 52.27  & 54.30  & 26.88  & 55.33  & 57.10  & 74.02  & 45.81  & 35.92   \\ 
Co-teaching & 47.23  & 38.44  & 668.22  & 235.00  & 60.50  & 54.30  & 23.44  & 65.03  & 61.66  & 75.06  & 45.97  & 32.31   \\ 
JoCor & 46.73  & 38.56  & 633.31  & 259.64  & 51.35  & 53.10  & 23.12  & 63.06  & 61.34  & 74.67  & 47.13  & 37.63   \\  \midrule
\multicolumn{12}{l}{\textit{Further fine-tuning}} \\
EL2N & 50.52  & 43.96  & 1046.46  & 261.43  & 52.98  & 57.30  & 26.42  & 64.95  & 64.81  & 74.62  & 47.11  & 38.62   \\ 
GradNorm & \underline{54.69}  & \underline{47.92}  & 1190.76  & 303.21  & \textbf{84.32}  & 56.20  & 26.65  & 65.98  & 64.91  & 75.36  & 48.60  & 34.20   \\ 
Entropy & 49.71  & 43.82  & 1159.33  & \underline{305.00}  & 42.42  & 57.20  & 27.11  & 61.86  & 63.09  & 72.43  & 45.07  & 37.72   \\ 
PPL & 54.68  & 45.52  & \underline{1305.64}  & 296.79  & 77.34  & 53.50  & 27.48  & \textbf{68.64}  & 64.91  & \underline{75.56}  & 47.68  & 38.45   \\ 
\rowcolor{backcolor}
Val\_PPL & \textbf{59.15}  & \textbf{52.02}  & \textbf{1417.26}  & \textbf{307.14}  & \underline{83.19}  & \textbf{66.50}  & \textbf{32.61}  & \underline{68.38}  & \textbf{66.22}  & \textbf{76.10}  & \textbf{52.02}  & \textbf{44.29}   \\ 
\bottomrule
\end{tabular}%
}
\caption{\textbf{Comparisons of different corruption-robust strategies at a corruption ratio of 50\% on LLaVA-1.5 (Qwen-2.5-7B).} Here, \textbf{Avg.} refers to the average performance on 11 benchmarks (normalized to 0-100). Best results are \textbf{Bold}, second best are \underline{underlined}.}
\label{tab:main_qwen_7b}
\end{table*}



\begin{table*}[!ht]
    \centering
    \begin{tabular}{lcccccc}
        \toprule
        \multirow{2}{*}{Models} & \multicolumn{2}{c}{EL2N} &  \multicolumn{2}{c}{GradNorm} &  \multicolumn{2}{c}{Entropy}    \\ 
        & $\downarrow$  & $\uparrow$ & $\downarrow$ & $\uparrow$ & $\downarrow$ & $\uparrow$  \\ \midrule
        Qwen-2.5-0.5B & \textbf{36.88}  & 28.11  & \textbf{47.07}  & 40.32  & \textbf{30.18}  & 31.67   \\ 
        Qwen-2.5-3B & \textbf{51.06}  & 30.41  & \textbf{49.49}  & 34.09  & \textbf{48.81}  & 38.01   \\ 
        Qwen-2.5-7B & \textbf{50.52}  & 23.14  & \textbf{54.69}  & 38.52  & \textbf{49.71}  & 39.98   \\ 
        LLaMA-3.1-8B & \textbf{47.07}  & 23.32  & \textbf{55.77} & 39.81  & \textbf{48.60}  & 30.73   \\  \bottomrule
    \end{tabular}
    \caption{\textbf{Further fine-tuning on the bottom ($\downarrow$) and top ($\uparrow$) 30\% subsets based on EL2N, GradNorm, and Entropy scores.} The reported results represent the average performance. All models are initially fine-tuned on data with a corruption ratio of $cr=50\%$.}
    \label{tab:ablation_lower_higher}
\end{table*}

\section{Extended Related Work on MLLMs} MLLMs integrate the vision modality into LLMs~\citep{touvron2023llama,chen2023gaining}, enabling the advanced understanding and reasoning over visual instructions~\citep{liu2023llava,bai2023qwen,gou2023mixture,gou2024eyes,chen2024emova}.
Recent VLLM works can be categorized into three directions, 1) \textit{Vision encoders}~\citep{oquab2023dinov2,chen2021multisiam,chen2023mixed} are enhanced and aggregated for robust representations~\citep{lin2023sphinx,li2024mini,tong2024cambrian}.
2) \textit{High-resolution} methods are proposed to overcome the fixed resolution of pre-trained vision encoders 
(\eg, $336 \times 336$ for CLIP~\citep{radford2021learningtransferablevisualmodels}), enabling LLMs to perceive fine-grained visual information~\citep{liu2024llavanext,dong2024xcomposer2-4khd,huang2024hires,luo2024feast}.
3) \textit{High-quality instruction data} is essential for VLLMs to generate accurate and well-formed responses~\citep{deitke2024molmo,chen2024expanding,li2025eagle}.
This paper is related to the third directions. However, rather than constructing high-quality data, we study study the effect of corrupted data on MLLMs and its mitigation. 
\end{document}